\documentclass{article}
\usepackage[margin=1in]{geometry}
\usepackage{graphicx}
\usepackage{amsmath}
\usepackage{hyperref}
\usepackage{graphicx}
\usepackage{amsmath, amssymb}
\usepackage{booktabs}
\usepackage{multirow}
\usepackage{hyperref}
\usepackage{titlesec}
\usepackage{caption}
\usepackage{subcaption}
\usepackage{listings}
\usepackage{xcolor}

\usepackage{tikz}
\usetikzlibrary{arrows.meta, positioning}

\lstdefinelanguage{Solidity}{
    keywords={contract, struct, event, function, returns, mapping, address,
              string, uint256, uint, bool, pure, view, external, internal,
              public, private, indexed, require, emit, return, pragma},
    keywordstyle=\color{blue}\bfseries,
    ndkeywords={uint256, string, address, bytes32},
    ndkeywordstyle=\color{teal},
    identifierstyle=\color{black},
    commentstyle=\color{gray}\itshape,
    stringstyle=\color{orange},
    sensitive=true,
    morecomment=[l]{//},
    morecomment=[s]{/*}{*/},
    morestring=[b]",
}

\usepackage{enumitem}
\usepackage{tikz}
\usetikzlibrary{positioning,arrows.meta,fit,shapes,calc}
\usepackage[
  style=ieee,   
  sorting=none  
]{biblatex}
\title{SlideChain: Semantic Provenance for Lecture Understanding via Blockchain Registration}

\author{
  Md Motaleb Hossen Manik \\
  Department of Computer Science \\
  Rensselaer Polytechnic Institute \\
  Troy, New York 12180, USA
  \and
  Md Zabirul Islam \\
  Department of Computer Science \\
  Rensselaer Polytechnic Institute \\
  Troy, New York 12180, USA
  \and
  Ge Wang \thanks{Corresponding author: Ge Wang, email: \texttt{wangg6@rpi.edu}} \\
  Department of Biomedical Engineering \\
  Rensselaer Polytechnic Institute \\
  Troy, New York 12180, USA
}

\date{}

\begin{document}
\maketitle

\begin{abstract}
Modern vision--language models (VLMs) are increasingly used to interpret and generate educational content, yet their semantic outputs remain challenging to verify, reproduce, and audit over time. Inconsistencies across model families, inference settings, and computing environments undermine the reliability of AI-generated instructional material, particularly in high-stakes and quantitative STEM domains. This work introduces \textbf{SlideChain}, a blockchain-backed provenance framework designed to provide verifiable integrity for multimodal semantic extraction at scale.
Using the SlideChain Slides Dataset—a curated corpus of 1,117 medical imaging lecture slides from a university course—we extract concepts and relational triples from four state-of-the-art VLMs and construct structured provenance records for every slide. SlideChain anchors cryptographic hashes of these records on a local EVM (Ethereum Virtual Machine)-compatible blockchain, providing tamper-evident auditability and persistent semantic baselines. Through the first systematic analysis of semantic disagreement, cross-model similarity, and lecture-level variability in multimodal educational content, we reveal pronounced cross-model discrepancies, including low concept overlap and near-zero agreement in relational triples on many slides.
We further evaluate gas usage, throughput, and scalability under simulated deployment conditions, and demonstrate perfect tamper detection along with deterministic reproducibility across independent extraction runs. Together, these results show that SlideChain provides a practical and scalable step toward trustworthy, verifiable multimodal educational pipelines, supporting long-term auditability, reproducibility, and integrity for AI-assisted instructional systems.
\end{abstract}


\section{Introduction}

Multimodal VLMs have rapidly advanced in their ability and are increasingly being used to interpret complex educational materials, including scientific figures, mathematical formulas, imaging geometries, and lecture slides. Recent surveys and benchmarks highlight progress in multimodal pretraining and downstream task performance \cite{du2022survey}, while large instruction-tuned VLMs such as LLaVA, Qwen-VL, and InternVL demonstrate impressive promises in captioning, question answering, knowledge extraction, and diagram interpretation. These capabilities position VLMs as potential tools for augmenting STEM instruction, enhancing lecture understanding, and enabling intelligent tutoring. Yet despite their rapid advancement, the \emph{semantic reliability} of modern VLMs remains poorly understood, particularly in a field-specific fashion, such as for biomedical imaging education.

A growing body of evidence shows that VLM outputs can be \emph{highly inconsistent} across model families, parameter scales, configurations, and even repeated runs on the same hardware. Surveys on hallucinations and semantic instability in large vision–language systems \cite{liu2024survey,sahoo2024comprehensive} document persistent issues such as ungrounded reasoning, missing or spurious concepts, and non-repeatable outputs. For a single input slide, different models may extract divergent concepts, produce incompatible relational triples, or emphasize different semantic structures. Such variability is problematic in educational and scientific settings where fidelity, precision, consistency, and interpretability are essential. Incomplete and unstable semantic interpretations risk distorting understanding, propagating misinformation, and undermining trust in AI-generated instructional material.

Complementing this challenge is the near absence of \emph{provenance and auditability mechanisms} in existing multimodal pipelines. As models evolve, datasets are updated, prompts change, or hardware environments shift, there is often no verifiable way to determine whether a semantic record reflects its authentic original state. Foundational work on data lineage, including the W3C PROV model and subsequent studies of scientific reproducibility, emphasizes the importance of transparent and traceable workflows. Yet recent analyses show that machine-learning pipelines are uniquely vulnerable to reproducibility failures: small perturbations in preprocessing, undocumented randomness, or data leakage can significantly alter downstream results \cite{semmelrock2023reproducibility,kapoor2022leakagereproducibilitycrisismlbased,beam2020challenges}. Importantly, these studies focus on reproducibility of \emph{models and datasets}, whereas in multimodal educational systems the primary object requiring verification is the \emph{semantic output} itself—concepts, relations, and extracted knowledge. No existing framework provides fine-grained, cryptographically verifiable provenance for these semantic structures.

To address these limitations, we introduce \textbf{SlideChain}, to the best of our knowledge the first blockchain-backed provenance and semantic analysis framework for multimodal educational slides. SlideChain integrates (i) a multimodal semantic extraction pipeline using four state-of-the-art VLMs namely InternVL3, Qwen2-VL, Qwen3-VL, and LLaVA-OneVision, (ii) a unified concept–triple schema enabling cross-model comparison, and (iii) a lightweight, EVM-compatible smart contract deployed on a local Hardhat chain to anchor cryptographic commitments of each slide’s provenance record. This design enables tamper-proof integrity checks, semantic drift detection, and long-term reproducibility audits. Using the \textbf{SlideChain Slides Dataset}— a curated corpus of 1,117 medical imaging slides from a 23-lecture university course—we conduct the first study of VLM semantic stability on real educational material and pair it with a comprehensive blockchain performance evaluation.

\paragraph{Contributions.}  
Our main contributions can be summarized as follows:

\begin{itemize}[leftmargin=12pt]
    \item \textbf{Multimodal semantic extraction pipeline.}  
    We develop a unified extraction framework that produces structured concepts, relational triples, and metadata for each slide in the SlideChain Slides Dataset using four state-of-the-art VLMs.

    \item \textbf{Blockchain-backed provenance system.}  
    We design and deploy \emph{SlideChain}, an efficient EVM-compatible smart contract that stores keccak256 commitments for all slide-level semantic records, providing immutable, verifiable provenance.

    \item \textbf{First stability analysis of VLMs on educational materials.}  
    Through cross-model disagreement analysis, triple sparsity evaluation, and pairwise Jaccard similarity measurements, we reveal substantial semantic incoherence and instability even on well-structured STEM instructional material.

    \item \textbf{Comprehensive blockchain performance evaluation.}  
    We measure gas usage, cost projection, block distribution, throughput, and scalability, demonstrating constant-gas behavior and approximately one-slide-per-second registration throughput.

    \item \textbf{Reproducibility and tamper-detection assessment.}  
    We show that the full semantic pipeline is deterministic across independent runs (Jaccard = 1.0 for all models) and that SlideChain achieves 100\% detection of adversarially perturbed provenance files.
\end{itemize}

Together, these contributions establish SlideChain as a practical, extensible, and principled foundation for reliable multimodal educational AI. Beyond quantifying its stability, the system provides a tamper-proof ledger for preserving, auditing, and reproducing multimodal semantic features over time—an essential requirement for scalable, transparent, and reliable AI-assisted STEM education.

The remainder of this paper is organized as follows.  
Section~\ref{sec:related-works} reviews related work.  
Section~\ref{sec:system-overview} presents the SlideChain system architecture.  
Section~\ref{sec:dataset-and-model} describes the dataset and vision--language models used in this study. 
Section~\ref{sec:semantic-analysis-methods} details the semantic extraction and analysis methodology.  
Section~\ref{sec:blockchain-methods} introduces the blockchain design and experimental setup.  
Section~\ref{sec:integrity-and-reproducibility-evaluation} evaluates integrity and reproducibility.  
Section~\ref{sec:results} reports empirical results.  
Section~\ref{sec:discussion} discusses implications and limitations.  
Finally, Sections~\ref{sec:conclusion} and~\ref{sec:future-works} conclude the paper and outline future research directions.

\section{Background and Related Work}
\label{sec:related-works}
This section situates SlideChain within three areas of prior work: multimodal slide understanding, provenance and reproducibility in AI systems, and blockchain mechanisms for scientific integrity. Together, these strands of literature highlight the need for verifiable, tamper-proof semantic provenance when applying modern VLMs to educational material.

\subsection{Multimodal Slide Understanding and VLM Limitations}

Recent progress in large vision--language models has dramatically expanded the ability of multimodal systems to interpret complex visual and textual inputs, including diagrams, figures, and structured document layouts. Models such as LLaVA~\cite{liu2023visualinstructiontuning}, Qwen-VL~\cite{bai2023qwenvlversatilevisionlanguagemodel}, and InternVL3~\cite{zhu2025internvl3exploringadvancedtraining} demonstrate strong capabilities in image-grounded reasoning, caption generation, and instruction-following tasks across diverse real-world settings. Despite these advances, their behavior on structured educational slides—where semantics combine visual layout, mathematical notation, and dense conceptual links—remains insufficiently studied.

Slide understanding has emerged as a distinct research area within multimodal AI. Prior works introduced datasets focusing on text detection, segmentation, or video–slide alignment, including SlideVQA~\cite{tanaka2023slidevqadatasetdocumentvisual}, educational multimodality datasets such as  Lecture Presentation Modality and Multimodal Multiview Audio-Visual, and lecture-driven multimodal corpora such as the ICCV Lecture Presentations dataset~\cite{lee2023lecture}. More recently, Zhang et~al.\ proposed the LecSlides-370K dataset~\cite{zhang2025towards}, a large-scale benchmark emphasizing global slide comprehension and question answering across diverse academic fields. These efforts highlight the growing importance of robust slide understanding but also reveal the challenges inherent to multimodal educational materials.

A persistent obstacle for current VLMs is semantic inconsistency. Even state-of-the-art models trained with sophisticated fusion modules frequently produce divergent interpretations for the same input slide. These inconsistencies manifest as missing or spurious concepts, contradictory relational structures, or mis-grounded explanations, a limitation documented both empirically in large-scale slide datasets~\cite{zhang2025towards} and theoretically in comparative evaluations of VLM behavior~\cite{bai2023qwenvlversatilevisionlanguagemodel}. InternVL3, for example, substantially improves visual grounding and multimodal alignment, yet still exhibits sensitivity to layout complexity and symbol-rich educational content~\cite{zhu2025internvl3exploringadvancedtraining}.

Beyond inconsistency, modern VLMs also suffer from hallucination, in which models produce semantically plausible but factually incorrect statements. Multiple surveys highlight hallucination as a fundamental limitation across language, image, and multimodal foundation models~\cite{liu2024surveyhallucinationlargevisionlanguage, sahoo-etal-2024-comprehensive}. These works document failures in grounding, overgeneralization, and misinterpretation—issues that become especially pronounced in scientific or diagrammatic content. In educational slides, hallucination manifests as invented relationships, incorrect formula explanations, or misidentification of imaging components, undermining downstream reliability.

Collectively, the literature demonstrates that while VLMs excel at general-purpose multimodal tasks, their semantic stability on domain-intensive educational slides remains limited. These systematic weaknesses motivate the need for verifiable provenance, multi-model comparison, and reproducibility frameworks—objectives directly addressed by the SlideChain system.

\subsection{Provenance and Reproducibility in AI Systems}

Ensuring reproducibility, verifiable lineage, and long-term auditability has become a central concern in modern machine learning. As models grow in scale and complexity, and as multimodal pipelines incorporate heterogeneous data sources, the ability to trace how results were produced—and to verify their fidelity over time—has become increasingly important. Prior work in scientific workflow provenance, dataset versioning, and reproducible ML systems highlights foundational challenges that motivate the design of frameworks such as SlideChain.

A major strand of provenance research builds on the PROV Data Model (PROV-DM), the W3C standard for representing entities, activities, and agents within scientific workflows~\cite{belhajjame2013prov}. PROV-DM offers a structured, graph-based approach to tracking data lineage and documenting transformations, and it has been widely adopted in scientific computing and workflow management systems. However, these frameworks generally assume deterministic transformations and do not address the instability of semantic outputs produced by modern neural models—particularly VLMs whose outputs may vary across architectures, inference settings, or hardware environments.

Reproducibility challenges in machine learning further underscore the need for robust provenance. Semmelrock et~al.~\cite{semmelrock2023reproducibilitymachinelearningdrivenresearch} survey how small, often undocumented pipeline variations can yield substantial drift in model outputs, inhibiting reliable scientific conclusions. Kapoor and Narayanan~\cite{kapoor2022leakagereproducibilitycrisismlbased} identify subtle data leakage as a major driver of irreproducible ML results, especially in published research. In the healthcare domain, Beam et~al.~\cite{beam2020challenges} highlight how model retraining, preprocessing inconsistencies, or hardware variation can significantly alter predictions, reinforcing the need for fine-grained tracking and versioning. These studies collectively demonstrate that traditional logs, code snapshots, or dataset versions are insufficient for ensuring semantic reproducibility, particularly in multimodal pipelines.

Modern ML engineering tools attempt to address these issues through dataset versioning and pipeline management. Systems such as DVC~\cite{Tuychiev2024DVC} provide structured mechanisms for tracking training data, configurations, and model artifacts, while production-focused guidelines emphasize explicit provenance and dependency capture for deployed ML services~\cite{machine-learning-in-production-ch24}. Yet these tools primarily track artifacts, not semantics. They do not provide guarantees that the conceptual or relational structures extracted by a vision–language model remain stable over time, nor can they detect semantic drift caused by model updates or nondeterministic inference.

Taken together, prior work establishes the importance of provenance, but existing tools operate at the level of data, code, and computational operations—not at the level of multimodal semantics. SlideChain extends this lineage by anchoring the semantic outputs themselves using cryptographic commitments. Related concerns regarding provenance, traceability, and accountability have also emerged in educational AI systems, where opaque model behavior can directly affect instructional content and learner understanding. By recording per-slide concept and triple extractions on an immutable ledger, SlideChain enables verifiable detection of semantic drift, ensures reproducibility across model runs, and provides long-term auditability for multimodal educational pipelines. This fills a critical gap in current reproducibility infrastructure: guaranteeing that the meaning extracted by an AI model remains traceable, inspectable, and unchanged.

\subsection{Blockchain for Scientific Integrity}

Ensuring long-term integrity of scientific and educational data has motivated increasing interest in blockchain-backed provenance systems. Early foundational work such as \textit{OpenTimestamps}~\cite{todd2016opentimestamps} demonstrated that blockchains can serve as decentralized, tamper-evident timestamping services, providing cryptographic guarantees of data existence at specific points in time. This approach established an important principle: storing only compact commitments (e.g., hashes) on-chain yields strong integrity without requiring full data storage on the ledger.

Building on these ideas, several systems proposed blockchain architectures explicitly designed for data provenance. ProvChain~\cite{liang2017provchain} showed that provenance metadata for cloud environments can be anchored on-chain to enhance privacy, availability, and tamper resistance. Follow-up work by Ruan et al.~\cite{ruan2019fine} introduced fine-grained, efficient provenance mechanisms for blockchain environments, enabling scalable and secure lineage tracking. More recently, BSTProv~\cite{sun2022bstprov} explored trustworthy provenance sharing using blockchain-backed verification, emphasizing secure dissemination and access control.

In scientific computing, SciChain~\cite{al2021scichain} demonstrated the feasibility of deploying blockchain-based provenance in high-performance computing (HPC) environments. The system introduced new consensus protocols tailored to resource-constrained scientific workflows and showed that blockchain can serve as a practical substrate for reproducible computational experiments. Complementing these system-level contributions, Akbarfam and Maleki’s Systematization of Knowledge survey~\cite{akbarfam2024sok} provides a comprehensive taxonomy of blockchain-for-provenance approaches, highlighting key design choices, security properties, and application domains.

Despite these advances, existing blockchain provenance efforts focus primarily on computational workflows, cloud systems, or general-purpose data integrity. None examine the unique challenges of multimodal semantic extraction or VLM-driven educational pipelines, where semantic drift, model nondeterminism, and long-term reproducibility demand fine-grained, per-artifact commitments. SlideChain fills this gap by applying blockchain principles to the semantic layer itself—anchoring concept and triple extractions at slide-level granularity to ensure tamper-proof, auditable provenance for multimodal educational content.

\section{System Overview}
\label{sec:system-overview}

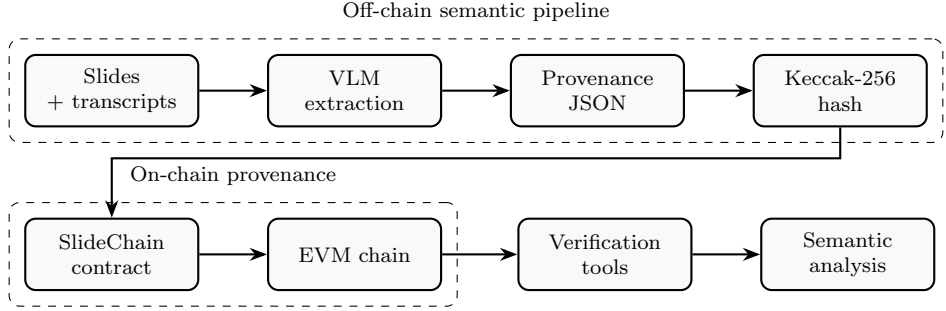
\begin{figure}[t]
\centering
\begin{tikzpicture}[
    font=\footnotesize,
    >=Stealth,
    line/.style={-Stealth,thick},
    block/.style={draw,rounded corners,thick,align=center,
                  minimum width=2.3cm,minimum height=0.95cm,fill=gray!5},
    group/.style={draw,rounded corners,dashed,inner sep=0.20cm},
    node distance=0.95cm
]

\node[block] (slides) {Slides\\+ transcripts};
\node[block,right=0.9cm of slides] (vlms) {VLM\\extraction};
\node[block,right=0.9cm of vlms] (prov) {Provenance\\JSON};
\node[block,right=0.9cm of prov] (hash) {Keccak-256\\hash};

\node[group,fit=(slides) (vlms) (prov) (hash),
      label={[align=center,above=0.1cm] {Off-chain semantic pipeline}}] (offchain) {};

\node[block,below=1.2cm of slides] (contract) {SlideChain\\contract};
\node[block,right=0.9cm of contract] (chain) {EVM chain};

\node[group,fit=(contract) (chain),
      label={[align=center,above=0.1cm] {On-chain provenance}}] (onchain) {};

\node[block,right=1.0cm of chain] (verify) {Verification\\tools};
\node[block,right=0.9cm of verify] (analysis) {Semantic\\analysis};

\draw[line] (slides) -- (vlms);
\draw[line] (vlms) -- (prov);
\draw[line] (prov) -- (hash);
\draw[line] (hash) |- ++(0,-0.9) -| (contract.north);
\draw[line] (contract) -- (chain);
\draw[line] (chain.east) -- ++(0.5,0) |- (verify.west);
\draw[line] (verify) -- (analysis);

\end{tikzpicture}

\caption{Three-layer system overview of the SlideChain framework. Layer 1: Off-chain, VLMs extract concepts and relational triples from slides and transcripts, which form provenance JSON records that are hashed. Layer 2: On-chain, the SlideChain contract stores immutable commitments. Layer 3: Verification tools and downstream semantic stability analysis.}
\label{fig:system_overview}
\end{figure}

SlideChain integrates multimodal semantic extraction with blockchain-backed provenance to create a verifiable, auditable, and reproducible pipeline for educational slide understanding. The framework is composed of four major layers: (i) slide ingestion and preprocessing, (ii) multimodel semantic extraction, (iii) provenance construction and hashing, and (iv) on-chain registration and verification. Figure~\ref{fig:system_overview} illustrates the end-to-end workflow.
\paragraph{Design Rationale.}
SlideChain is explicitly designed to provide provenance for \emph{semantic outputs}, rather than for models or datasets alone. In educational and scientific settings, trust and reproducibility depend on what semantic interpretation was produced for a given artifact, not merely which model version existed at a given time. Because modern vision--language models often produce divergent interpretations for the same input, a single-model pipeline can obscure semantic instability and create a false sense of determinism. SlideChain therefore adopts a multi-model extraction strategy to surface, rather than suppress, semantic variability across state-of-the-art VLMs. From a systems perspective, the framework records only cryptographic commitments on-chain while keeping full semantic records off-chain: hashing ensures immutability and verifiability with minimal cost, while off-chain storage preserves flexibility, scalability, and rich semantic structure. This hash-on-chain / semantics-off-chain design represents a minimal yet sufficient provenance mechanism—any weaker design fails to guarantee auditability, while stronger on-chain storage is unnecessary for integrity verification.

\subsection{Pipeline Summary}

The SlideChain pipeline begins with the ingestion of high-resolution slide images and accompanying transcript snippets from the \textbf{SlideChain Slides Dataset}, a curated collection of 1,117 educational slides from a medical imaging course. Each slide is processed independently through four state-of-the-art VLMs, each configured for multimodal concept and relational triple extraction. The resulting outputs differ widely in structure and completeness, necessitating a normalization stage.

All raw model outputs are converted into a unified provenance JSON file that captures concepts, triples, evidence fragments, and contextual metadata. This representation provides a structured semantic footprint for each slide and serves as the basis for downstream stability analysis.

To ensure long-term auditability, SlideChain computes a Keccak-256 hash for every provenance JSON file. These hashes act as compact, immutable commitments to the underlying semantic content. Each hash is then recorded on a \textbf{local EVM-compatible Hardhat blockchain} using the \texttt{registerSlide} function of the SlideChain smart contract. The resulting on-chain entry binds every $(\text{lecture\_id}, \text{slide\_id})$ pair to its cryptographic commitment, an optional URI, and a timestamp.

Verification tools allow users—or automated downstream systems—to recompute a slide’s provenance hash locally and compare it against the on-chain value via the \texttt{isRegistered} and \texttt{getSlide} interfaces. This enables fine-grained integrity checks, drift detection, version auditing, and reproducibility verification, fully independent of model version, inference hardware, or software environment.

\subsection{Knowledge Representation}

Semantic outputs from the four VLMs are transformed into a unified, model-agnostic representation consisting of three core components:

\paragraph{Concepts.}  
Atomic semantic units extracted from each slide, encoded as \texttt{(category, term)} pairs. Categories span \emph{modality}, \emph{anatomy}, \emph{workflow}, \emph{software}, and other domain-specific labels. Concepts may include short evidence snippets, linking extracted semantics back to their textual or visual origins.

\paragraph{Triples.}  
Structured relational assertions of the form $(s, p, o)$, where $s$ denotes a subject entity, $p$ a predicate phrase, and $o$ an object entity. Triples provide higher-level relational structure and capture causal, descriptive, hierarchical, or compositional relationships often present in scientific slides.

\paragraph{Metadata.}  
Each provenance record includes slide identifiers, model names, extraction timestamps, input paths, and all raw evidence returned by the VLMs. Metadata ensures interpretability years after extraction and enables deterministic reproducibility checks across model versions and hardware configurations.

All components are organized within a hierarchical JSON schema designed to accommodate missing fields, inconsistent formatting, or model-specific deviations. This schema forms the foundation for SlideChain’s semantic disagreement metrics, Jaccard similarity analysis, and reproducibility evaluation.
\subsection{SlideChain Smart Contract}

At the core of the SlideChain provenance layer is a lightweight smart contract written in Solidity and deployed on a \textbf{local Hardhat EVM-compatible development chain}. The contract serves as an immutable registry that binds each educational slide to a cryptographic commitment of its semantic provenance, forming the backbone of verifiable semantic lineage.

For every $(\text{lecture\_id}, \text{slide\_id})$ pair, the contract stores a compact provenance record consisting of three elements: (i) a Keccak-256 hash of the canonical slide-level provenance JSON, (ii) an optional URI pointing to the off-chain location of the full semantic record, and (iii) the block timestamp at which the slide was registered. Together, these fields provide a minimal yet sufficient on-chain representation that enables integrity verification, temporal ordering, and external retrieval without exposing large semantic payloads on-chain.

Slide registration is performed through the \texttt{registerSlide} function, which records the cryptographic commitment and associated metadata for a given slide. To preserve immutability, the contract explicitly rejects duplicate registrations for the same slide identifier. Because the function performs only constant-time storage operations with no loops or dynamic memory allocation, it exhibits predictable and near-constant gas usage across all registrations. This design choice is critical for scalability, ensuring that provenance costs remain stable as the dataset grows.

Once registered, provenance records can be retrieved using the \texttt{getSlide} interface, which returns the stored hash, URI, timestamp, and registrant address for a specified slide. A complementary \texttt{isRegistered} function enables efficient existence checks without retrieving full metadata, supporting lightweight verification workflows in downstream tools.

The contract is intentionally minimalist by design. It prioritizes predictable gas consumption, scalability to large slide collections, and compatibility with Ethereum-like execution environments, while storing only cryptographic commitments on-chain. All semantic content remains off-chain, preserving flexibility in storage, format evolution, and access control. This separation allows SlideChain to provide strong integrity guarantees and reproducibility support without incurring the prohibitive costs of on-chain semantic storage.

Overall, this smart contract architecture enables verifiable, tamper-evident provenance for multimodal educational semantics while remaining efficient, portable, and suitable for both experimental evaluation and future deployment on public EVM-compatible networks.

\section{Dataset and Models}
\label{sec:dataset-and-model}
This work uses the \textbf{SlideChain Slides Dataset} as the semantic substrate for evaluating multimodel extraction, cross-model variability, and blockchain-backed provenance. The dataset consists of educational slides from a full-semester medical imaging course, and four state-of-the-art VLMs are applied to every slide to generate complementary concept and relation extractions that form the basis of our semantic and provenance analyses.
\subsection{SlideChain Slides Dataset}

The SlideChain Slides Dataset consists of \textbf{23 full-length university lectures} in medical imaging, spanning core topics such as X-ray physics, computed tomography reconstruction, magnetic resonance imaging principles, ultrasound imaging, PET/SPECT, and multidimensional signal processing. From these lectures, we curated a total of \textbf{1,117 high-resolution slide images}, each paired with a refined transcript segment extracted from the corresponding lecture narration.

The dataset exhibits substantial semantic and visual diversity. Slides range from diagrammatic depictions of imaging geometry and physical principles, to mathematical equations and multi-step derivations, system block diagrams and reconstruction workflows, clinical imaging examples, and text-dense conceptual summaries. This heterogeneity reflects realistic instructional material used in advanced STEM courses and presents a nontrivial challenge for multimodal semantic extraction.

Such diversity makes the dataset particularly well suited for evaluating semantic consistency and stability across vision--language models. Different slides emphasize visual reasoning, mathematical structure, or linguistic explanation to varying degrees, allowing us to probe how model behavior shifts across content types. Each slide is assigned a unique identifier of the form $(\text{lecture\_id}, \text{slide\_id})$ and is processed independently through the semantic extraction and provenance pipeline described in the following sections.

\subsection{Vision--Language Models}

We evaluate four modern VLMs that collectively span major design paradigms in multimodal AI, including transformer-based architectures, strong visual encoders, instruction tuning, and cross-modal alignment. These models differ significantly in capacity, training methodology, and output density, providing a broad characterization of semantic behavior.

\paragraph{InternVL3–14B.}  
A 14B-parameter multimodal transformer with advanced spatial grounding and dense cross-modal alignment. InternVL3 typically produces the richest semantic representations, excelling in fine-grained concept identification and contextual extraction.

\paragraph{Qwen2–VL–7B.}  
A high-capacity vision--language model optimized for general-purpose reasoning. Qwen2–VL–7B generates diverse concept sets and handles both diagrammatic and text-heavy slides effectively.

\paragraph{Qwen3–VL–4B.}  
A lightweight, instruction-tuned VLM with efficient cross-modal fusion. Although smaller in scale, it produces structured concept lists suitable for downstream comparison, with a more conservative semantic footprint.

\paragraph{LLaVA-OneVision (Qwen2–7B backbone).}  
An instruction-following multimodal model designed for conversational visual understanding. While effective in qualitative explanation tasks, it generates sparser structured outputs, providing a useful contrast to more analytically oriented models.

Together, these four VLMs reveal a wide spectrum of semantic extraction behavior—from dense to sparse concept production, and from consistent to highly variable relational output.

\subsection{Semantic Extraction Pipeline}

Each slide undergoes a unified four-step extraction pipeline that integrates both visual and textual cues. For every VLM and every slide, the following operations are performed:

\begin{enumerate}
    \item \textbf{Input construction:}  
    Each slide image is paired with its corresponding transcript snippet to form a multimodal input prompt (image + text).

    \item \textbf{Model inference:}  
    The VLM outputs structured information, including \emph{concepts}, \emph{triples}, \emph{evidence snippets}, and optional \emph{free-form explanatory text}. Output structure and completeness vary substantially across models.

    \item \textbf{Normalization:}  
    All raw model outputs are parsed into a consistent schema:
    \[
        \texttt{provenance} = \{ \text{concepts},\; \text{triples},\; \text{evidence},\; \text{metadata} \}.
    \]
    This step performs null-safe parsing, canonicalizes text, removes duplicates, and tolerates the heterogeneous formats emitted by different VLMs.

    \item \textbf{Provenance JSON generation:}  
    Each slide’s multimodel semantic record is stored in a machine-verifiable JSON file. This file serves as the canonical artifact for blockchain hashing, semantic disagreement analysis, and reproducibility evaluation.
\end{enumerate}

An illustrative excerpt of the unified JSON schema is shown below where we include both fetched and missing triples.

\begin{verbatim}
{
  "lecture": "Lecture 1",
  "models": {
    "InternVL3-14B": {
      "concepts": [
        {"category": "modality", "term": "medical imaging"},
        {"category": "workflow", "term": "core principles"}
      ],
      "triples": [
        {"s": "Medical imaging", "p": "uses",
         "o": "science and engineering", "confidence": 1.0}
      ]
    },
    "Qwen3-VL-4B": {
      "concepts": [
        {"category": "physics", "term": "medicine"}
      ],
      "triples": []
    }
  },
  "paths": {
    "image": ".../Lecture1/Images/Slide1.jpg",
    "text":  ".../Lecture1/Texts/Slide1.txt"
  }
}
\end{verbatim}

This unified representation forms the foundation for SlideChain’s subsequent semantic disagreement metrics, cross-model Jaccard similarity analysis, reproducibility testing, and blockchain-backed provenance registration.

\section{Semantic Analysis Methods}
\label{sec:semantic-analysis-methods}
To evaluate the reliability of multimodal semantic extraction, we conduct quantitative analyses over the concepts, relational triples, and evidence returned by four state-of-the-art VLMs. This section details the extraction, normalization, and statistical procedures used to characterize semantic disagreement, cross-model similarity, lecture-level trends, and stability categories across the SlideChain Slides Dataset.

\subsection{Normalization of Concepts and Triples}

Each VLM produces heterogeneous semantic outputs that may include structured lists, nested dictionaries, free-form text, or malformed fields. To support consistent downstream computation, SlideChain applies a robust normalization pipeline with the following steps:

\begin{enumerate}
    \item \textbf{Null-safe parsing:}  
    Concept sets may appear as lists, single objects, or \texttt{null}. All representations are harmonized into a flat list of \((\text{category}, \text{term})\) pairs, discarding empty or malformed entries.

    \item \textbf{Triple flattening:}  
    Triples sometimes contain extraneous metadata or nested attributes. We extract only the canonical relational tuple \((s, p, o)\), with optional confidence scores retained when available. Missing or malformed triple fields are treated as empty sets.

    \item \textbf{Evidence preservation:}  
    Textual evidence returned by VLMs is aggregated verbatim. Although not used in quantitative scoring, evidence is preserved for interpretability, provenance inspection, and future qualitative analysis.

    \item \textbf{Stable canonicalization:}  
    All concepts and triples are normalized via lowercasing, whitespace regularization, and deduplication. This removes superficial formatting inconsistencies across models.
\end{enumerate}

After normalization, each slide \(s\) and model \(M_i\) yields two finite sets:
\[
C_i(s) \quad \text{(concept set)}, \qquad 
T_i(s) \quad \text{(triple set)}.
\]

These consistent semantic representations form the foundation for all downstream metrics.

\subsection{Disagreement Metrics}

Semantic disagreement measures how widely the four VLMs diverge when interpreting the same slide. For each slide \(s\), we compute:

\[
D_{\text{concept}}(s)
= \left| \bigcup_{i=1}^{4} C_i(s) \right|,
\qquad
D_{\text{triple}}(s)
= \left| \bigcup_{i=1}^{4} T_i(s) \right|.
\]

The unions capture the \emph{total semantic breadth} proposed by any model. High disagreement values indicate that models extract substantially different concepts or relational structures—direct evidence of semantic instability.

Crucially, no model is treated as the ground truth. These metrics quantify the dispersion of outputs across the ensemble rather than comparing against a privileged reference model.

\subsection{Cross-Model Jaccard Similarity}

While disagreement measures global dispersion, we also compute \emph{pairwise} semantic alignment using the Jaccard coefficient. For any two models \(M_a\) and \(M_b\) and for any semantic set \(S \in \{C, T\}\), we compute:

\[
J(S_a, S_b)
=
\frac{|S_a \cap S_b|}
     {|S_a \cup S_b|}.
\]

A value of \(1.0\) denotes perfect agreement; a value of \(0\) indicates complete divergence.

We compute Jaccard similarity:
\begin{itemize}
    \item for every slide (per-slide similarity),
    \item for each model pair (averaged across all slides),
    \item separately for concepts and triples.
\end{itemize}

The resulting similarity matrices reveal which models align semantically and which diverge systematically.

\subsection{Lecture-Level Aggregation}

To study semantic patterns at the instructional level, we aggregate disagreement across each lecture. For a lecture \(L\) containing slides \(S_L\), we compute:

\[
\overline{D}_{\text{concept}}(L)
=
\frac{1}{|S_L|}
\sum_{s \in S_L} D_{\text{concept}}(s),
\qquad
\overline{D}_{\text{triple}}(L)
=
\frac{1}{|S_L|}
\sum_{s \in S_L} D_{\text{triple}}(s).
\]

These lecture-level scores reveal semantic variability trends across the instructional sequence. High-disagreement lectures commonly correspond to visually dense slides, multi-stage imaging pipelines, or mathematically intensive content.

\subsection{Semantic Stability Classification}

To identify which slides yield reliable semantic representations across models, we classify slides into stability categories using empirical percentiles of the disagreement distribution.

Let \(Q_1\) and \(Q_3\) denote the 25th and 75th percentiles of the concept disagreement values. Then:

\begin{itemize}
    \item \textbf{Stable:} \(D_{\text{concept}}(s) \le Q_1\)
    \item \textbf{Moderate:} \(Q_1 < D_{\text{concept}}(s) \le Q_3\)
    \item \textbf{Unstable:} \(D_{\text{concept}}(s) > Q_3\)
\end{itemize}

Triple disagreement may also be incorporated into composite stability metrics, but concept disagreement dominates due to the sparsity and higher variance of triple extraction.

This classification provides a principled mechanism to locate semantically robust slides, identify ambiguous or model-sensitive content, and guide downstream provenance inspection and quality assurance in multimodal educational pipelines.

\section{Blockchain Methods}
\label{sec:blockchain-methods}
SlideChain implements on-chain provenance through an EVM-compatible smart contract deployed on a deterministic Hardhat development network. This section describes the execution setup, registration workflow, and measurement methodology used to evaluate the practicality, cost, and scalability of blockchain-backed provenance for multimodal educational datasets.
\subsection{Experimental Setup}

All blockchain and semantic extraction experiments were conducted on a controlled local workstation. Running both components on fixed hardware and software configurations ensures full reproducibility of the SlideChain pipeline. The hardware and software specifications are presented in \ref{tab:hardware} and \ref{tab:software}, respectively.

\subsubsection*{Hardware Environment}

Experiments were performed on a multi-GPU workstation. While semantic extraction utilizes GPU parallelism, all blockchain measurements rely solely on CPU execution and local disk I/O.

\begin{table}[h]
\centering
\caption{Hardware specifications used for SlideChain experiments.}
\begin{tabular}{p{4cm} p{7.5cm}}
\toprule
\textbf{Component} & \textbf{Details} \\
\midrule
GPUs & 4 $\times$ NVIDIA RTX A5000 (24 GB VRAM each) \\
CPU & Intel Xeon-class processor \\
System RAM & 256 GB \\
CUDA version & 12.1 \\
Storage & Local SSD (blockchain + provenance artifacts) \\
\bottomrule
\end{tabular}
\label{tab:hardware}
\end{table}

\subsubsection*{Software Environment}

Semantic extraction and analysis were executed in Python, while blockchain evaluation was performed in a Node.js-based Hardhat environment.

\begin{table}[!ht]
\centering
\caption{Software environment used for semantic extraction and blockchain evaluation.}
\begin{tabular}{p{4cm} p{7.5cm}}
\toprule
\textbf{Software Component} & \textbf{Version / Description} \\
\midrule
Python & 3.10 \\
PyTorch & 2.2 (CUDA 12.1) \\
Transformers & HuggingFace official implementations for all VLMs \\
Scientific Stack & NumPy, Pandas, Matplotlib, Seaborn \\
Node.js & v22.10.0 \\
Hardhat & v3.0.17 \\
Solidity Compiler & 0.8.20 \\
Operating System & Windows 11 \\
\bottomrule
\end{tabular}
\label{tab:software}
\end{table}
\subsubsection*{Determinism and Reproducibility}

All preprocessing routines, prompts, and configuration parameters were held fixed across runs. Although GPU operations can introduce low-level nondeterminism, the complete semantic extraction pipeline produced bit-identical provenance JSONs across two full executions, yielding Jaccard~=~1.0 for all model--slide combinations. The behavior of the Hardhat development chain further contributes to reproducibility: the miner seals exactly one block per transaction with zero-delay mining, assigns stable block timestamps, uses deterministic EIP-1559 fee calculations, and preserves a fixed transaction ordering. Together, these properties ensure consistent gas measurements, predictable temporal behavior, and reliable verification outcomes across repeated experiments. We emphasize that this determinism reflects strict engineering control over prompts, model versions, decoding settings, and execution environment, rather than a general property of vision–language model inference.

\subsubsection*{Execution Scope}

All provenance JSON files, blockchain logs, gas metrics, and analysis outputs were generated entirely on local hardware. The SlideChain smart contract was deployed exclusively on a Hardhat development chain, providing full control over execution semantics and eliminating dependence on external validators or network variability. This controlled environment guarantees predictable mining behavior and reproducible gas consumption, forming a stable basis for the semantic, blockchain, and reproducibility analyses reported in Section~\ref{sec:results}. We note that all blockchain experiments in this study are conducted on a deterministic local Hardhat development chain to ensure reproducibility and controlled measurement of gas usage and throughput. In real-world EVM deployments, additional factors such as network latency, transaction reordering, temporary chain reorganizations, and gas-price volatility may affect registration timing and operational cost. However, these factors do not alter the cryptographic integrity guarantees of SlideChain: any post hoc modification of off-chain provenance records would still be detected through hash mismatch against on-chain commitments. Consequently, while deployment environment influences performance and cost characteristics, it does not affect the correctness or security of provenance verification.

\subsection{Local Hardhat Chain Setup}

To emulate Ethereum-style execution while ensuring full reproducibility, all blockchain operations were performed on a deterministic Hardhat development chain.  
Table~\ref{tab:hardhat-setup} summarizes the configuration.

\begin{table}[h]
\centering
\caption{Configuration of the local Hardhat development chain.}
\label{tab:hardhat-setup}
\begin{tabular}{p{4.2cm} p{7.8cm}}
\toprule
\textbf{Component} & \textbf{Configuration} \\
\midrule
RPC endpoint & \texttt{http://127.0.0.1:8545} \\
Development accounts & 20 pre-funded deterministic accounts (fixed private keys) \\
Gas pricing model & EIP-1559 base fee with deterministic priority tip \\
Block policy & One block sealed per transaction (zero-delay mining) \\
Contract deployment & \texttt{hardhat-ethers} \\
Invocation interface & Python scripts using exported ABI and contract address \\
Tracing & Full access to receipts, internal call traces, and gas metadata \\
\bottomrule
\end{tabular}
\end{table}

Hardhat's deterministic execution semantics ensure stable block numbering, reproducible timestamps, and consistent gas-cost behavior.  
This controlled environment enables precise measurement of registration throughput, cost scaling, and integrity properties across the full SlideChain pipeline.

\subsection{Slide Registration Procedure}

Each slide in the SlideChain Slides Dataset corresponds to a canonical provenance JSON file containing multimodel semantic information. To preserve long-term integrity, only the cryptographic hash of this file is written on-chain.

For each lecture identifier \(L\) and slide number \(s\), the provenance JSON is serialized with lexicographically sorted keys to ensure platform-invariant hashing, and the hash

\[
h_{L,s} = \text{keccak256}(\texttt{JSON}(L,s))
\]

is computed. Registration consists of:

\begin{enumerate}
    \item computing the Keccak-256 hash of the slide’s provenance JSON,
    \item constructing a canonical local URI for off-chain storage,
    \item invoking \texttt{registerSlide(lectureId, slideId, hash, uri)} on-chain,
    \item recording gas usage, timestamps, and block metadata from the receipt.
\end{enumerate}

All 1,117 slides were registered exactly once, producing 1,117 corresponding blockchain transactions.  
Each transaction emits a \texttt{SlideRegistered} event containing the stored hash and timestamp, forming a tamper-proof audit log for the entire extraction pipeline.

\subsection{Gas and Cost Measurement}

For every registration transaction, we extract:

\begin{itemize}
    \item \textbf{gasUsed}: execution gas consumed by \texttt{registerSlide},
    \item \textbf{effectiveGasPrice}: post-EIP-1559 gas price determined by Hardhat,
    \item \textbf{blockNumber}: used for analyzing block density and ordering.
\end{itemize}

The monetary cost of storing a provenance hash is computed as

\[
\text{cost}_{L,s} = \text{gasUsed}_{L,s} \times \text{effectiveGasPrice}_{L,s},
\]

and ETH-to-USD conversion uses a reference rate of \$3000/ETH.  
We report minimum, maximum, and mean gas per registration, total gas across all slides, and total cost in both ETH and USD.  
These measurements quantify the operational cost of integrating blockchain-backed provenance into educational AI systems.

\subsection{Block Distribution and Throughput}

Hardhat's “one block per transaction” mining policy ensures that each registration corresponds to a unique block.  
Let \(t_i\) denote the timestamp of transaction \(i\).  
The total registration throughput is

\[
\text{throughput} 
= \frac{1117}{t_{\text{last}} - t_{\text{first}}}
\approx 1.0009~\text{slides/sec},
\]

given that \(t_{\text{last}} - t_{\text{first}} = 1116\,\text{seconds}\).

This deterministic mining behavior produces a strictly increasing block height that aligns linearly with slide index, simplifying analysis of temporal patterns, block–slide mapping, and cumulative gas evolution.

\subsection{Scalability Simulations}

We extrapolate the cost and gas requirements for datasets ranging from \(10^3\) to \(10^6\) slides.  
For a dataset of size \(N\), total gas is

\[
G_{\text{total}} = N \times \overline{g},
\]

where \(\overline{g}\) is the empirical mean gas per registration, and total expected cost is  
\(G_{\text{total}} \times p_{\text{gas}}\) for a target network gas price \(p_{\text{gas}}\).  
Expected registration time is estimated using the empirical throughput of 1 slide/sec.

We evaluate three representative EVM environments:

\begin{itemize}
    \item \textbf{Ethereum L1 (30 gwei)}: highest decentralization and security, highest cost,
    \item \textbf{Polygon PoS (5 gwei)}: low-cost sidechain optimized for high-throughput applications,
    \item \textbf{Optimistic L2 (1 gwei)}: minimal-cost environment suitable for large-scale provenance storage.
\end{itemize}

Results show linear scaling with dataset size and highlight that L2 and sidechain networks can reduce total costs by one to two orders of magnitude, making million-slide provenance feasible with modest computational and financial overhead.

\section{Integrity and Reproducibility Evaluation Protocol}
\label{sec:integrity-and-reproducibility-evaluation}

To establish trust in multimodal semantic provenance, a provenance system must satisfy three core properties:  
(i) it must preserve a reliable temporal ordering of semantic records,  
(ii) it must detect any post hoc modification of provenance artifacts, and  
(iii) it must support reproducible semantic extraction under controlled conditions.  

This section describes the evaluation protocols used to assess these properties for SlideChain across the full set of 1,117 slides in the SlideChain Slides Dataset. Quantitative results and visualizations are presented in Section~\ref{sec:results}.

\subsection{Time-Gap Evaluation Protocol}

To assess temporal integrity, we compare the local filesystem modification time (\texttt{mtime}) of each provenance JSON file with the blockchain timestamp recorded when the corresponding slide is registered on-chain via \texttt{registerSlide}. For each slide with lecture identifier \(L\) and slide index \(s\), we define the time gap as

\[
\Delta t_{L,s} = t_{\text{chain}}(L,s) - t_{\text{local}}(L,s),
\]

where \(t_{\text{local}}\) denotes the provenance file creation time and \(t_{\text{chain}}\) the block timestamp at registration.

This metric captures the end-to-end delay between off-chain semantic generation and on-chain commitment. Because all provenance records are generated prior to blockchain registration, \(\Delta t\) reflects queuing, execution, and confirmation behavior under the local Hardhat mining policy. The distribution of \(\Delta t\) provides a basis for verifying consistent temporal ordering and detecting anomalous registration behavior.

\subsection{Tamper Detection Protocol}

To evaluate integrity guarantees, we test whether SlideChain can reliably detect corruption of off-chain provenance records. A subset of slides is selected at random, and controlled perturbations are applied to their provenance JSON files. These perturbations include modifications to concept terms, deletion or alteration of relational triples, injection of spurious semantic elements, and edits to evidence strings.

For each modified file, we recompute its Keccak-256 hash and compare it against the hash stored on-chain. Verification succeeds only if the recomputed hash matches the on-chain commitment. This procedure directly tests the system’s ability to detect unauthorized modifications to semantic records after registration.

Because the blockchain stores only cryptographic commitments, this evaluation isolates the effectiveness of hash-based integrity checking independent of storage medium or file format.

\subsection{Reproducibility Across Runs}

To assess semantic reproducibility, we perform two fully independent executions of the semantic extraction pipeline under identical prompts, model versions, and execution settings. Each run generates a complete set of provenance records covering all slides and all four VLMs, resulting in paired provenance files for every model--slide combination.

For each pair, we compute Jaccard similarity over extracted concept sets and relational triple sets:

\[
J_{\text{concept}} = \frac{|C_1 \cap C_2|}{|C_1 \cup C_2|},
\qquad
J_{\text{triple}} = \frac{|T_1 \cap T_2|}{|T_1 \cup T_2|},
\]

where \(C_1, C_2\) and \(T_1, T_2\) denote concepts and triples from the two runs, respectively.

These metrics quantify the degree of semantic consistency across independent executions and form the basis for detecting nondeterminism or drift under controlled conditions. When combined with on-chain commitments, this protocol enables future extractions to be audited against an immutable semantic baseline. 
We note that this observed determinism reflects strict engineering control over prompts, model versions, decoding settings, and execution environment, rather than a general property of vision--language model inference.

\section{Results}
\label{sec:results}

This section presents the empirical evaluation of SlideChain across three complementary dimensions.  
First, we analyze the semantic behavior and variability of four VLMs on 1{,}117 educational slides, including disagreement, cross-model similarity, lecture-level patterns, and stability characteristics.  
Second, we perform a lightweight baseline comparison between single-model and multi-model semantic provenance to quantify how much semantic content is missed when relying on a single VLM.  
Third, we evaluate the performance, scalability, and reliability of the blockchain-backed provenance layer, including gas consumption, throughput, cost projections, integrity guarantees, tamper detection, and reproducibility across independent extraction runs.  
All results are reported using the full SlideChain Slides Dataset.

\subsection{Semantic Model Analysis}

To assess the reliability and variability of multimodal semantic extraction, we evaluated 4{,}468 model--slide combinations (1,117 slides $\times$ 4 VLMs): InternVL3--14B, Qwen2--VL--7B, Qwen3--VL--4B, and LLaVA-OneVision--Qwen2--7B.  
Our analysis examines five aspects of semantic behavior: (i) disagreement in extracted concepts and triples, (ii) per-model semantic footprint, (iii) cross-model similarity, (iv) lecture-level variability, and (v) stability classification.  
Together, these analyses constitute the first quantitative assessment of how modern VLMs interpret university-level STEM slides.

\subsubsection{Concept and Triple Disagreement}

\begin{figure}[t]
    \centering
    \includegraphics[width=0.85\linewidth]{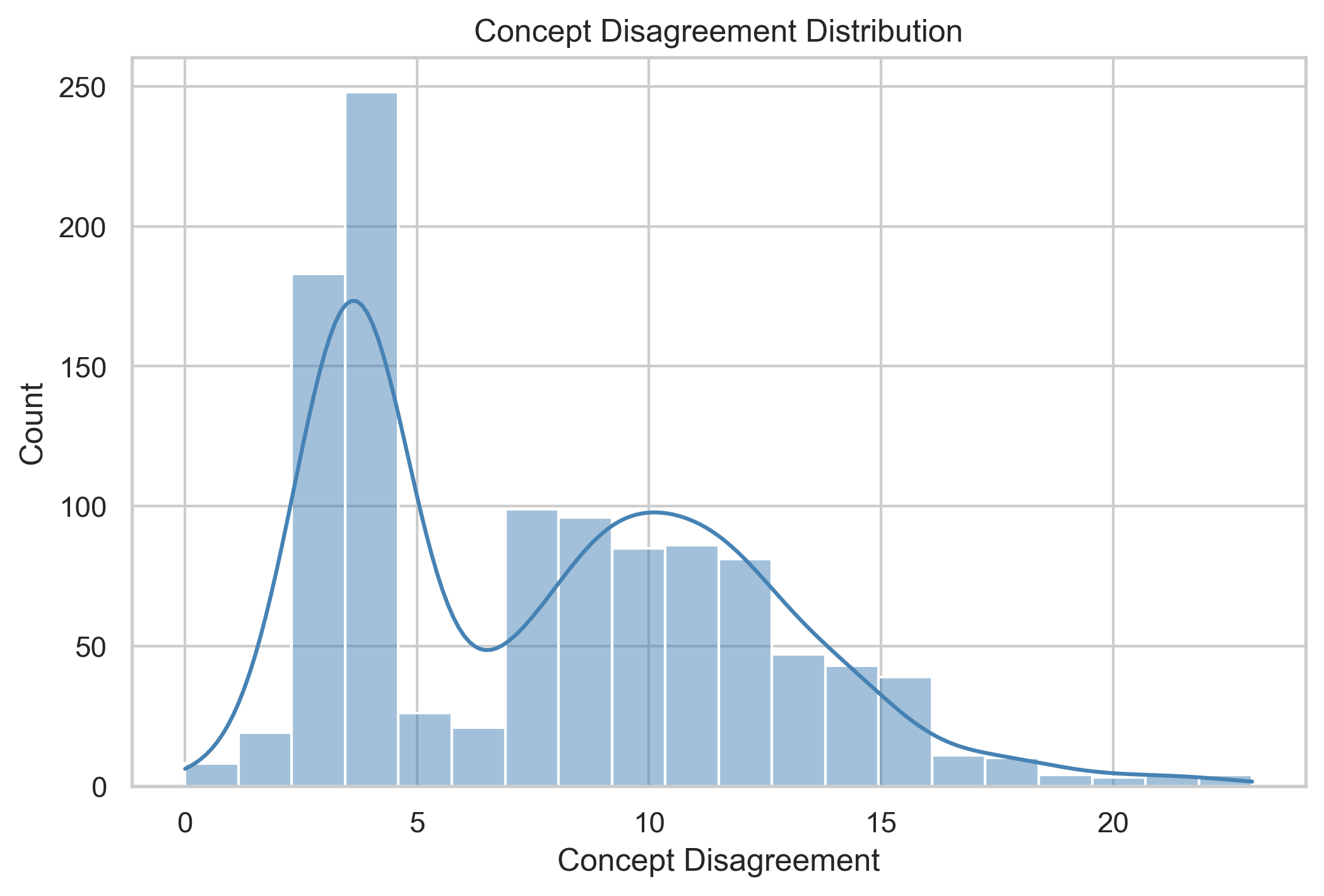}
    \caption{Distribution of concept disagreement across 1,117 slides.}
    \label{fig:concept-disagreement-hist}
\end{figure}

\begin{figure}[t]
    \centering
    \includegraphics[width=0.85\linewidth]{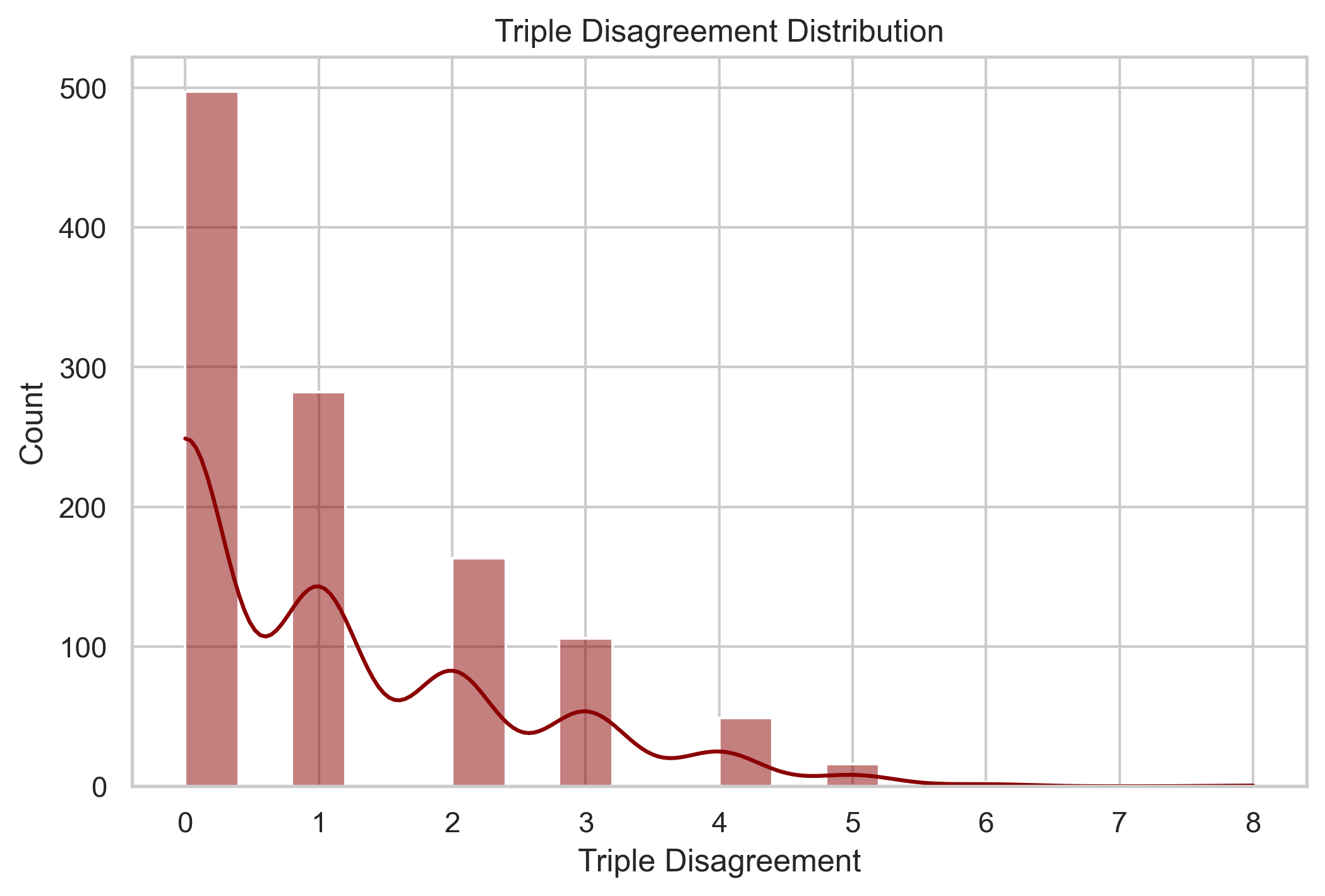}
    \caption{Distribution of triple disagreement across slides.}
    \label{fig:triple-disagreement-hist}
\end{figure}

Figures~\ref{fig:concept-disagreement-hist} and \ref{fig:triple-disagreement-hist} show the distributions of concept and triple disagreement.  
Concept disagreement frequently exceeds 8--12 unique terms per slide, indicating that different VLMs often extract markedly different concepts even when given identical multimodal input.

Triple disagreement is lower in magnitude because relational extraction is sparse overall, but still highly inconsistent across models.  
In many cases, triples proposed by one VLM do not appear in any other model’s output.  
This demonstrates a key limitation of current multimodal systems: while entity-level recognition is relatively robust, relational grounding—determining how entities interact or relate—remains fragile and model-dependent.

\subsubsection{Per-Model Concept and Triple Production}

\begin{figure}[t]
    \centering
    \includegraphics[width=0.85\linewidth]{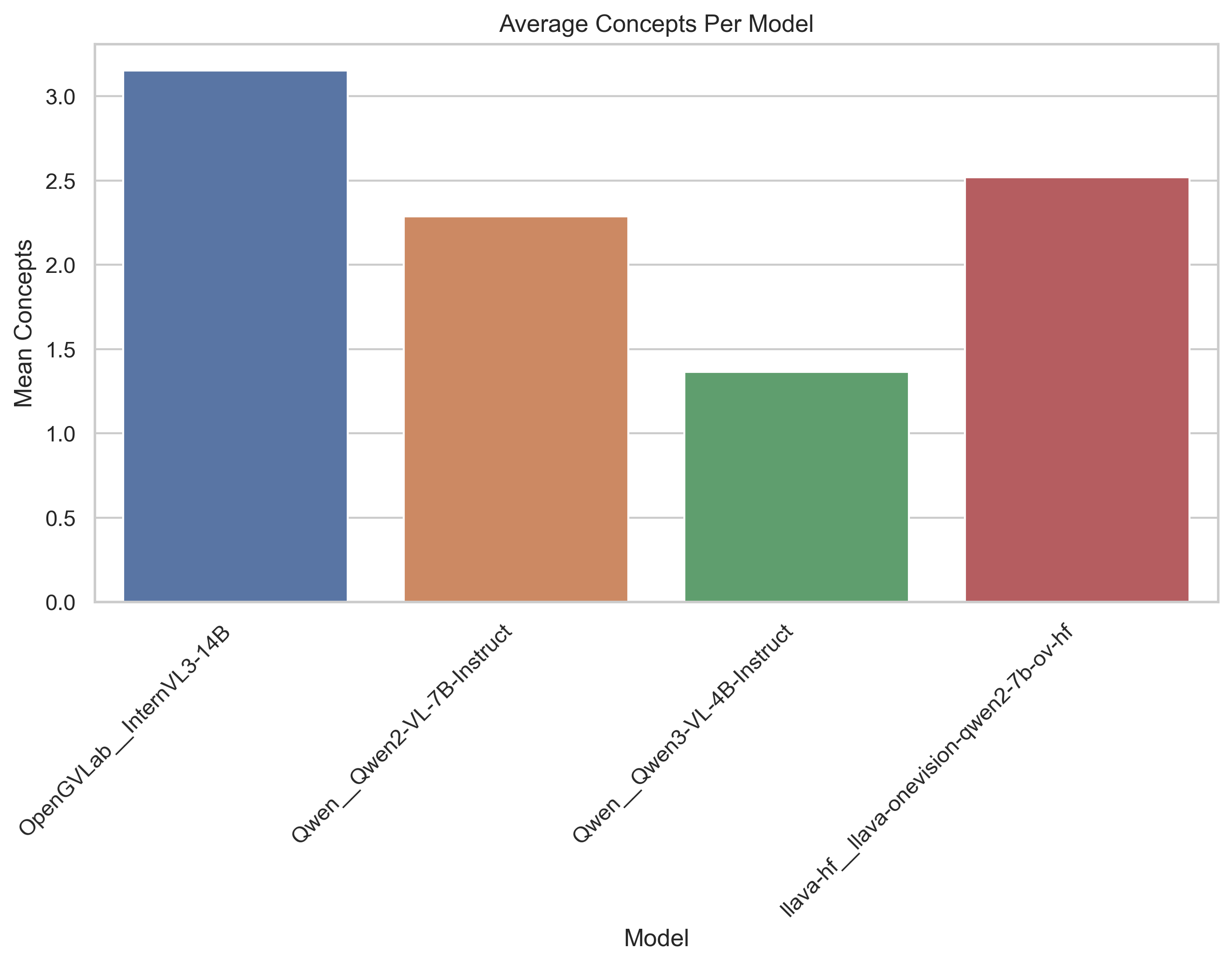}
    \caption{Average number of concepts produced per model.}
    \label{fig:model-concepts-mean}
\end{figure}

\begin{figure}[t]
    \centering
    \includegraphics[width=0.85\linewidth]{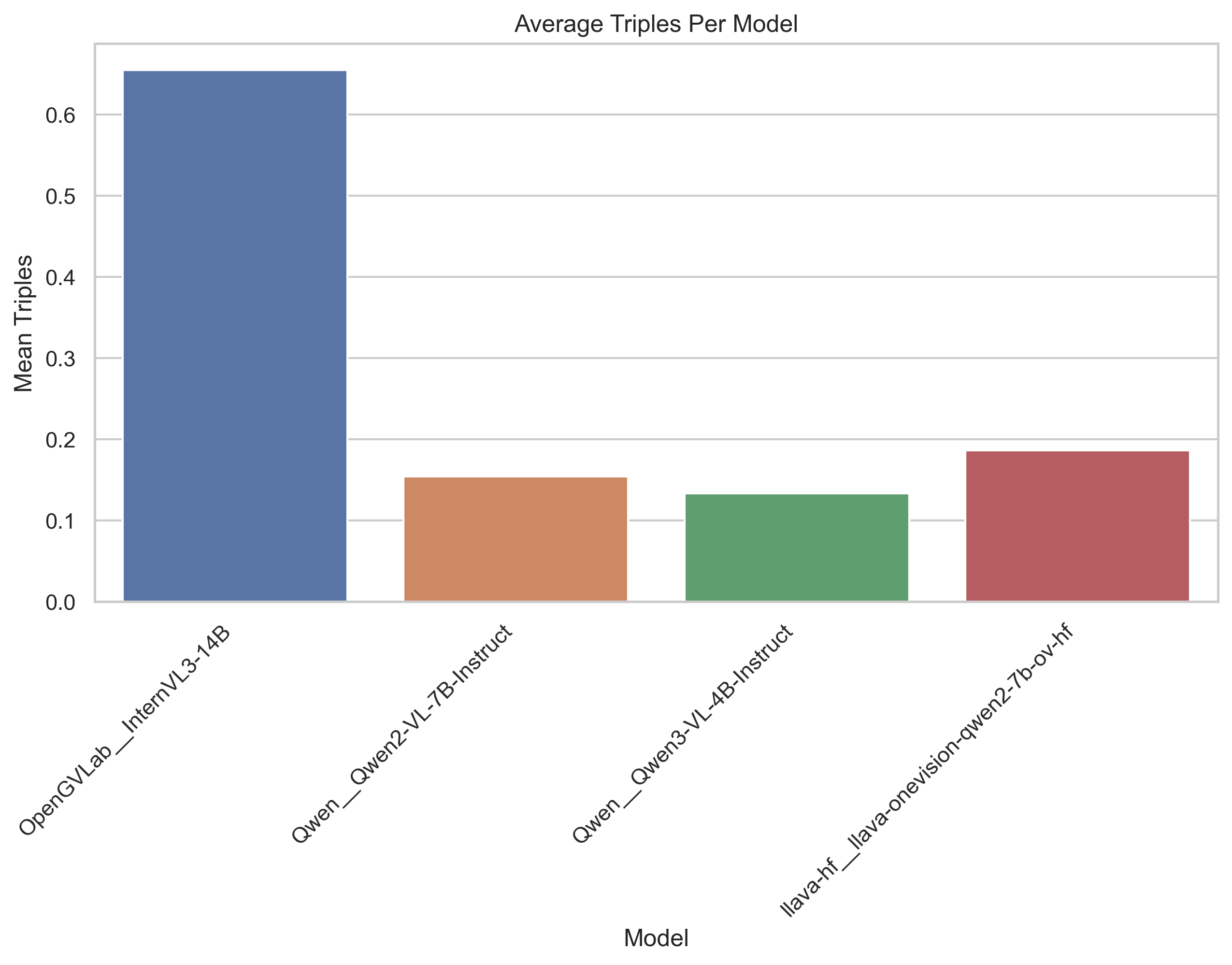}
    \caption{Average number of triples extracted per model.}
    \label{fig:model-triples-mean}
\end{figure}

To better understand the sources of disagreement, Figures~\ref{fig:model-concepts-mean} and~\ref{fig:model-triples-mean} show the average number of concepts and triples generated by each model.  
The models differ substantially in output density. InternVL3--14B produces the richest semantic representations, approximately twice as many concepts as Qwen3--VL--4B.  
Qwen2--VL--7B follows closely, while LLaVA-OneVision exhibits moderate concept extraction but consistently low triple production.

These differences highlight that VLMs not only disagree semantically but also operate at different levels of semantic granularity.  
Dense models contribute disproportionately to disagreement by capturing broader semantic footprints, while smaller models may omit important information entirely.  
Such heterogeneity underscores the need for provenance tracking: downstream systems must know \emph{which model} produced which interpretation and under what conditions.

\subsubsection{Cross-Model Semantic Similarity}

\begin{figure}[t]
    \centering
    \includegraphics[width=0.8\linewidth]{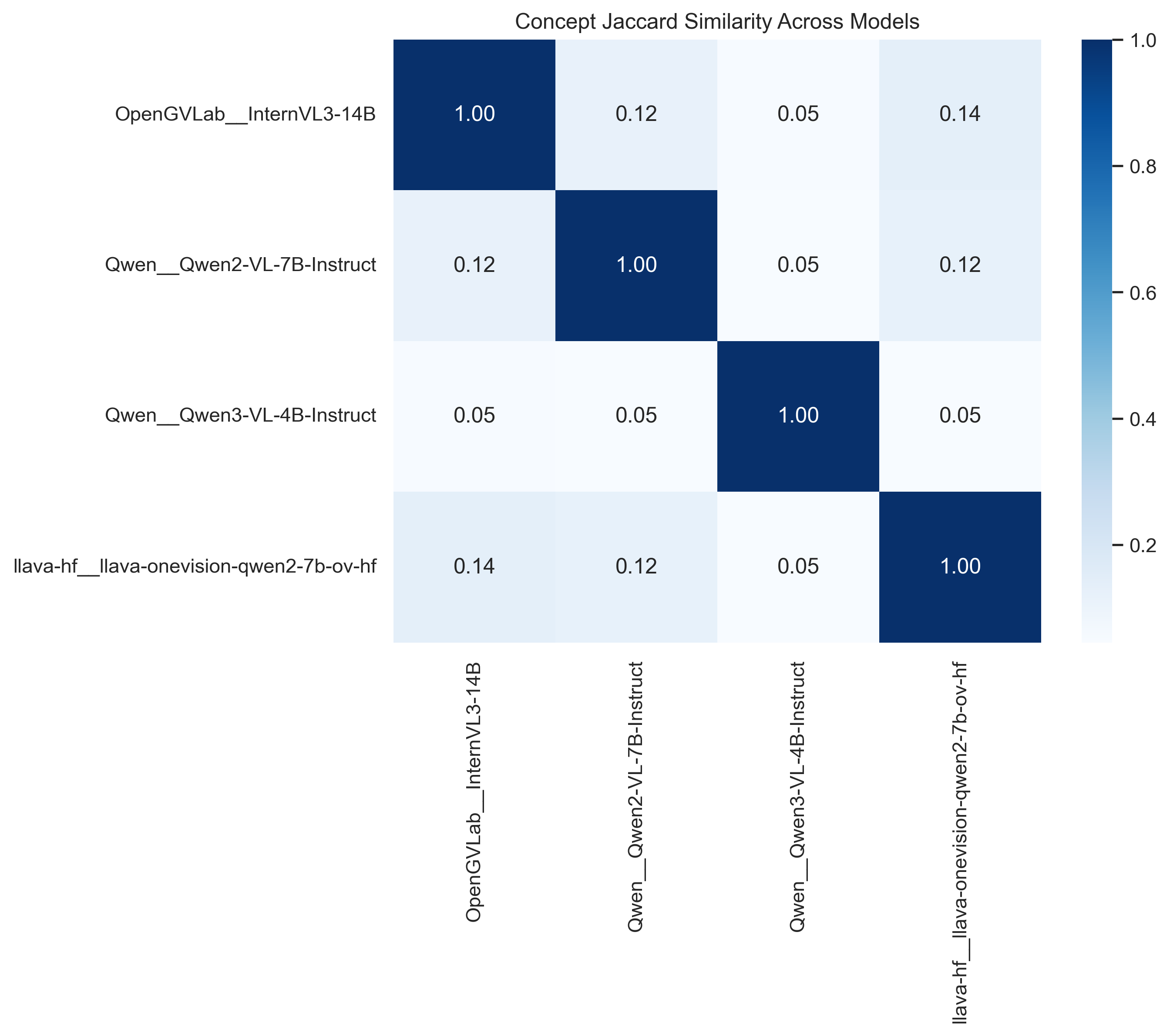}
    \caption{Pairwise Jaccard similarity for concept sets across models, showing consistently low cross-model semantic agreement despite identical inputs.
}
    \label{fig:concept-jaccard}
\end{figure}

\begin{figure}[t]
    \centering
    \includegraphics[width=0.8\linewidth]{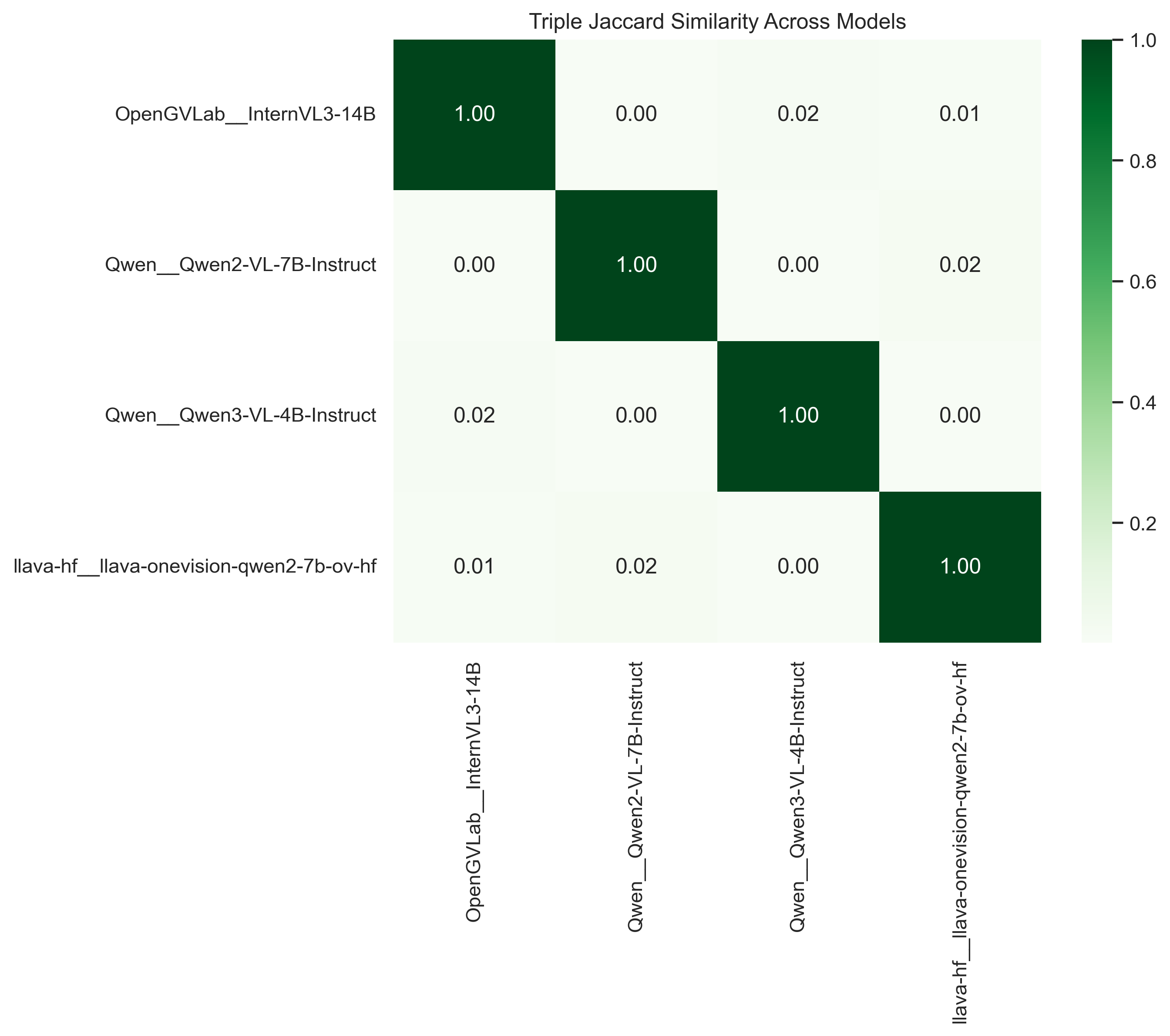}
    \caption{Pairwise Jaccard similarity for triples across models.}
    \label{fig:triple-jaccard}
\end{figure}

We compute pairwise Jaccard similarity for all model pairs to quantify semantic alignment beyond simple output counts.  
Figures~\ref{fig:concept-jaccard} and~\ref{fig:triple-jaccard} show the resulting similarity matrices for concepts and triples.

Concept overlap is consistently modest, demonstrating that even high-level interpretations vary substantially across architectures and training regimes.  
Triple overlap is near zero for several model pairs, reaffirming the difficulty VLMs face in reliably extracting structured relational information.  
These results reinforce the need for multi-model evaluation: no single VLM provides a comprehensive or authoritative semantic representation.

\subsubsection{Lecture-Level Disagreement}

\begin{figure}[t]
    \centering
    \includegraphics[width=0.85\linewidth]{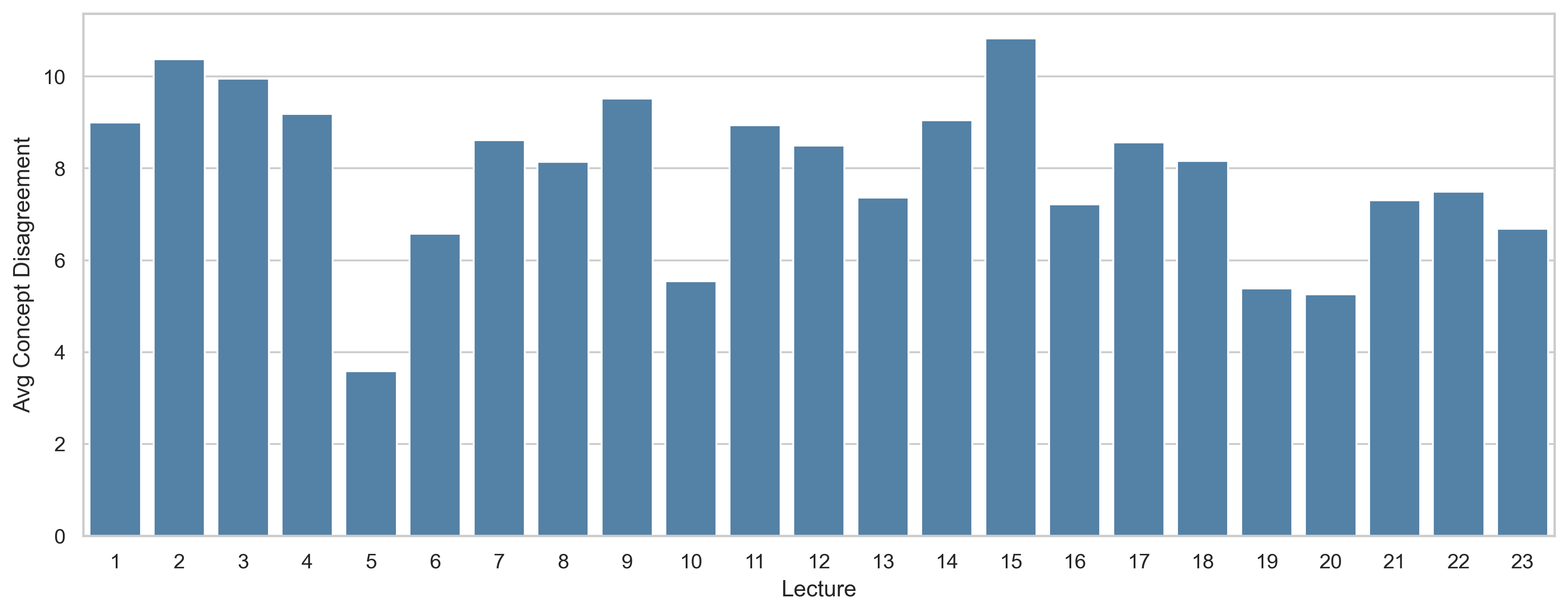}
    \caption{Average concept disagreement per lecture.}
    \label{fig:lecture-concept-disagreement}
\end{figure}

\begin{figure}[t]
    \centering
    \includegraphics[width=0.85\linewidth]{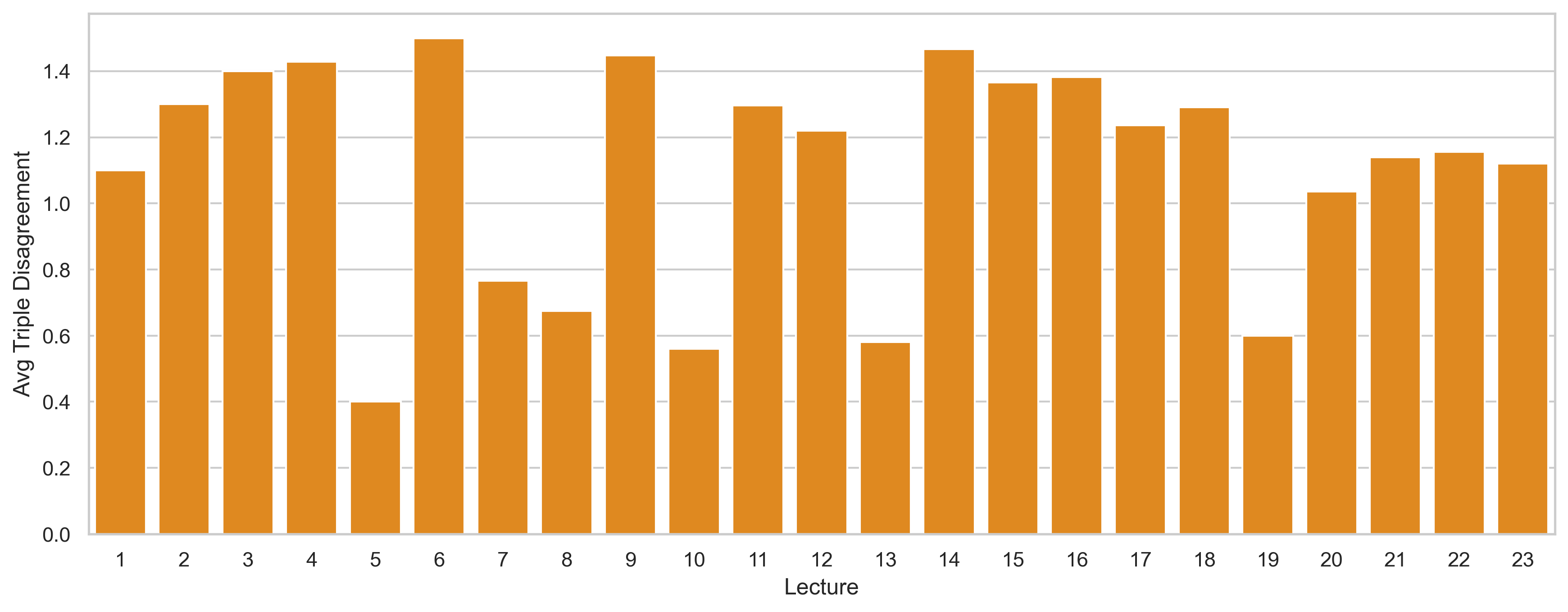}
    \caption{Average triple disagreement per lecture.}
    \label{fig:lecture-triple-disagreement}
\end{figure}

To examine whether semantic variability correlates with instructional content, we aggregate disagreement across lectures (Figures~\ref{fig:lecture-concept-disagreement} and~\ref{fig:lecture-triple-disagreement}).  
Lectures containing dense diagrams, physics-driven imaging pipelines, or multi-panel visualizations consistently exhibit higher disagreement, while text-dominant or conceptually simple slides show lower variability.

This trend aligns with pedagogical intuition: the most visually or conceptually dense slides are those where learners typically require additional guidance, and where VLM interpretations diverge most.  
Provenance-aware educational systems could leverage these patterns to identify and prioritize challenging instructional material.

\subsubsection{Semantic Stability Classification}

\begin{figure}[t]
    \centering
    \includegraphics[width=0.7\linewidth]{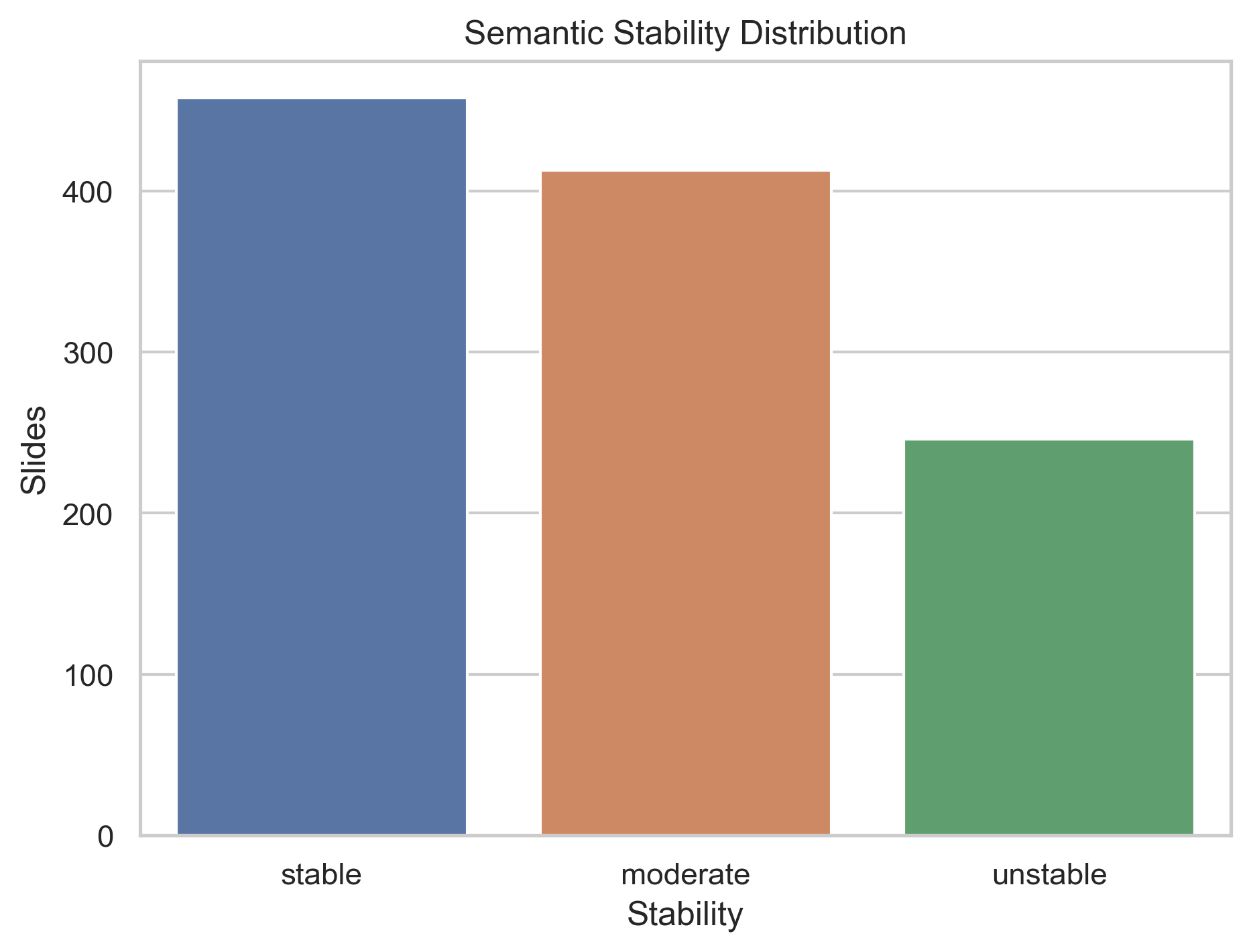}
    \caption{Distribution of semantic stability categories across all slides.}
    \label{fig:semantic-stability}
\end{figure}

Finally, we classify each slide as \emph{stable}, \emph{moderate}, or \emph{unstable} based on percentiles of concept disagreement.  
Figure~\ref{fig:semantic-stability} reports the distribution across all slides.

A large fraction of slides fall into the moderate or unstable ranges, indicating that scientific and engineering slides pose significant interpretation challenges even for advanced VLMs.  
This result emphasizes the necessity of verifiable provenance: without it, downstream systems cannot distinguish between semantic disagreements caused by content complexity, model limitations, or drift in the extraction pipeline.

\begin{figure}[t]
\centering
\includegraphics[width=0.75\linewidth]{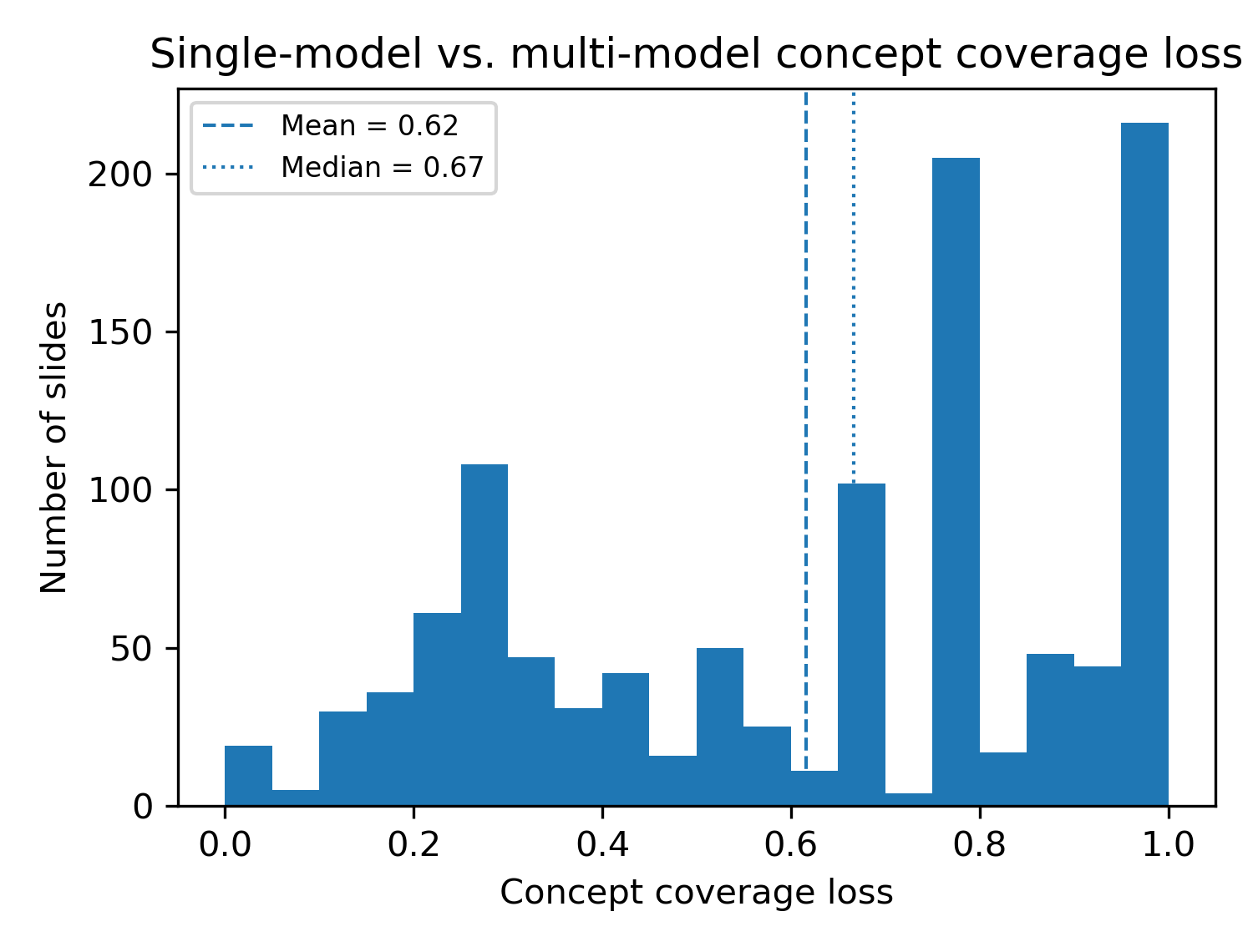}
\caption{Concept coverage loss when using a single VLM (InternVL3) compared to multi-model provenance. Values close to 1 indicate that most concepts present in the multi-model union are missed by the single-model baseline.}
\label{fig:concept-coverage-loss}
\end{figure}

\begin{figure}[t]
\centering
\includegraphics[width=0.75\linewidth]{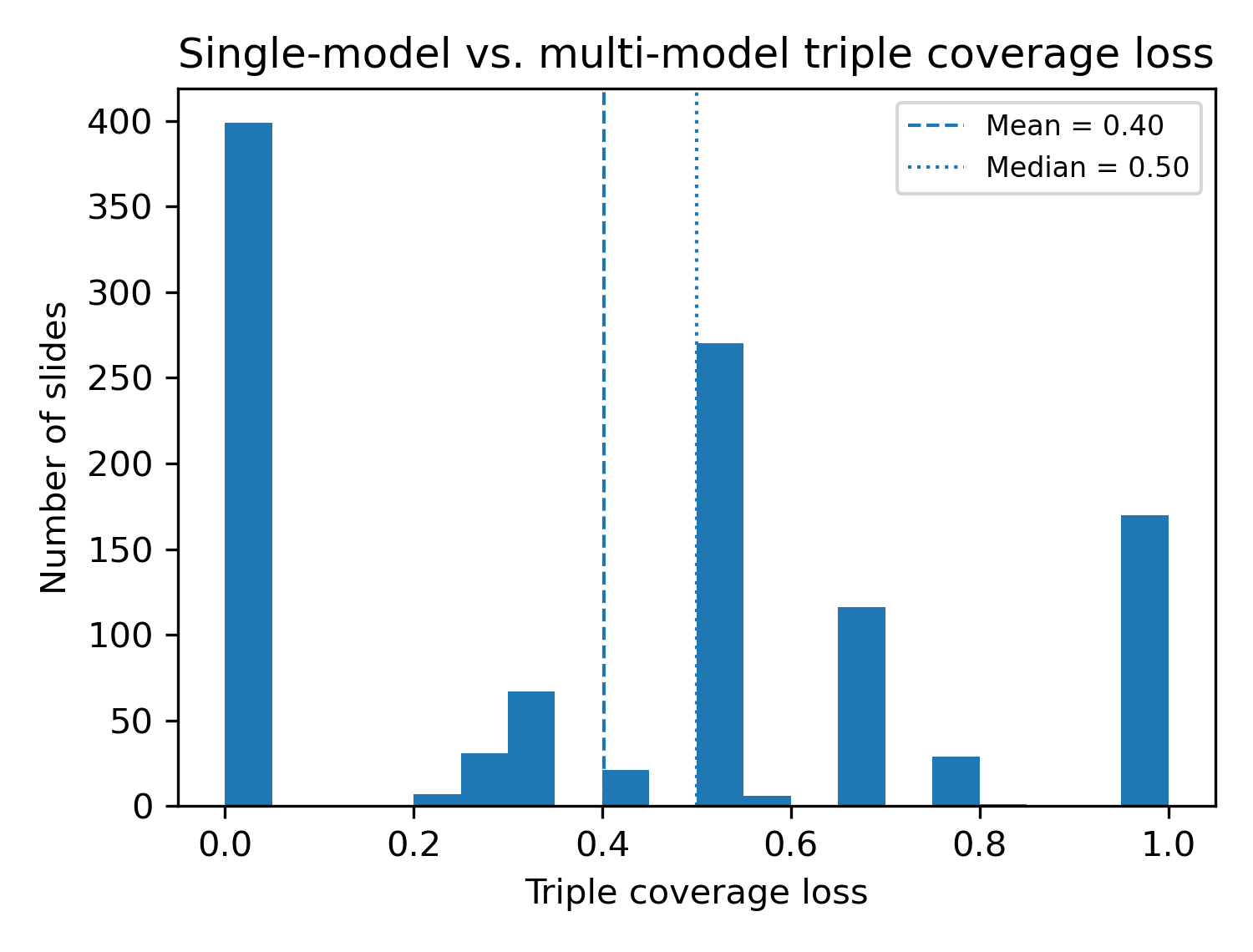}
\caption{Triple coverage loss under single-model provenance relative to multi-model extraction. Many slides exhibit partial or complete loss of relational semantics when relying on a single VLM.}
\label{fig:triple-coverage-loss}
\end{figure}

\subsection{Single-Model vs.\ Multi-Model Provenance Coverage}
\label{sec:single-vs-multi}

To motivate the need for multi-model semantic provenance, we perform a lightweight baseline comparison between single-model and multi-model extraction using existing outputs. We treat InternVL3 as a representative strong VLM and compare its extracted semantics against the union of concepts and triples obtained from all four models. InternVL3 is selected as the single-model baseline because it consistently produces the densest semantic footprint across slides, making the single-model comparison conservative with respect to coverage loss.

For each slide, we compute the fraction of semantic elements present in the multi-model union but absent from the single-model baseline. Figures~\ref{fig:concept-coverage-loss} and~\ref{fig:triple-coverage-loss} show the resulting coverage-loss distributions across all 1,117 slides.

On average, single-model provenance misses approximately 62\% of concepts identified by the multi-model pipeline, with a median loss of 67\%. Relational triples exhibit a mean coverage loss of 40\%, with many slides losing half or more of their extracted relations under the single-model baseline. In extreme cases, single-model extraction captures none of the relations present in the multi-model union.

These results indicate that relying on a single VLM substantially under-represents the semantic content of educational slides. Importantly, this variability is invisible without a multi-model comparison and motivates SlideChain’s design choice to treat semantic outputs—rather than individual models—as the object of provenance.

\subsection{Blockchain Registration Performance}

We evaluate the SlideChain smart contract under full-scale registration of all 1,117 slide-level provenance records.  
Our analysis spans gas consumption stability, block distribution, temporal throughput, cost characteristics, scalability projections, and on-chain integrity guarantees.  
These results collectively demonstrate that blockchain-backed provenance can be integrated into multimodal educational pipelines without prohibitive computational or financial cost.

\subsubsection{Gas Consumption Behavior}

\begin{figure}[t]
\centering
\includegraphics[width=0.8\linewidth]{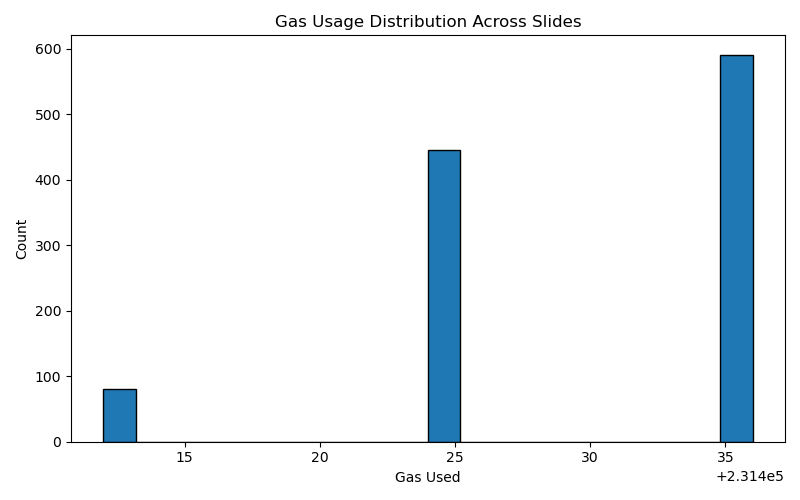}
\caption{Histogram of gas usage across all 1,117 registrations.}
\label{fig:gas-usage-hist}
\end{figure}

\begin{figure}[t]
\centering
\includegraphics[width=0.8\linewidth]{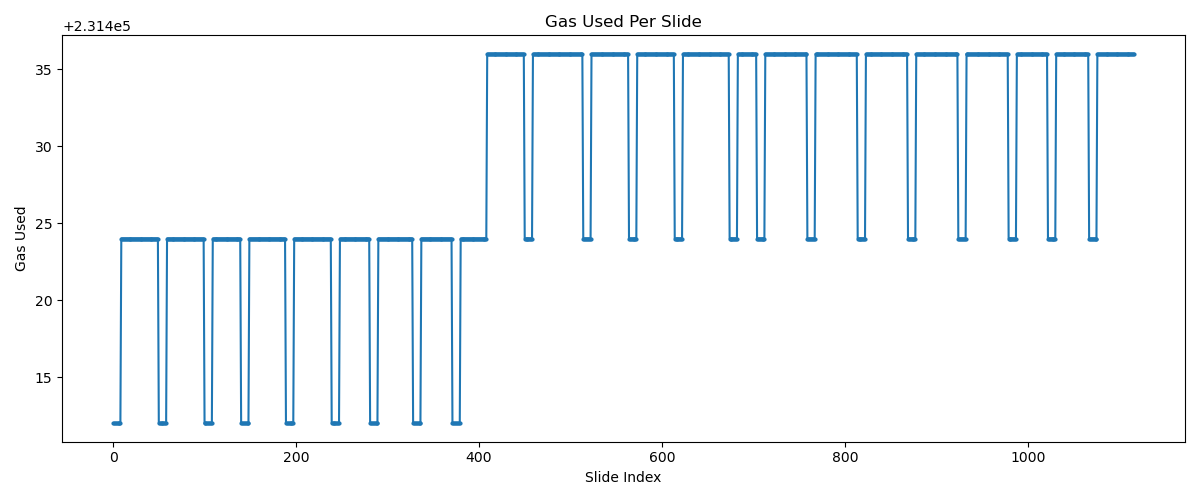}
\caption{Gas used per slide in sequential registration.}
\label{fig:gas-per-slide}
\end{figure}

\begin{figure}[t]
\centering
\includegraphics[width=0.8\linewidth]{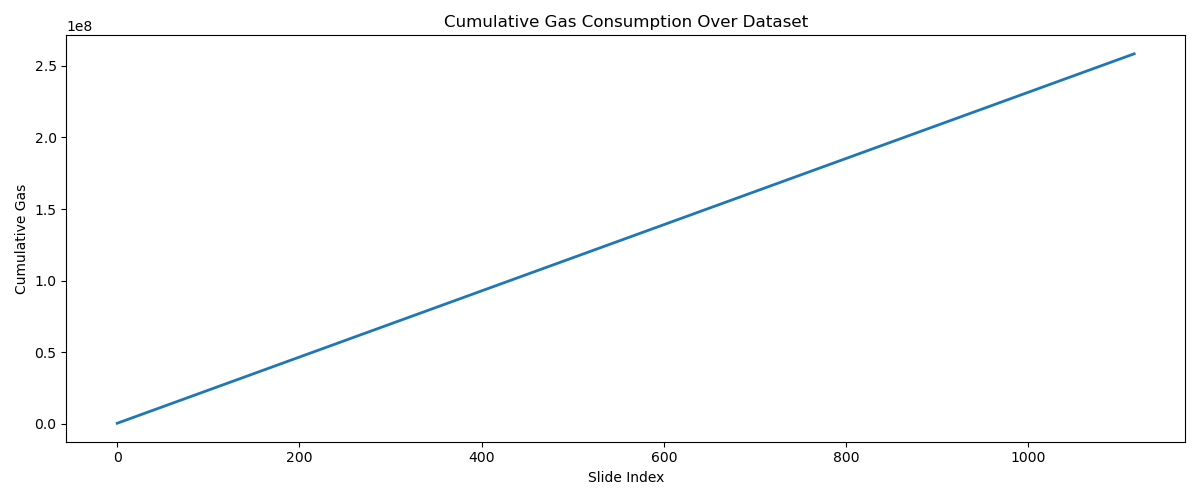}
\caption{Cumulative gas consumption over time.}
\label{fig:cumulative-gas}
\end{figure}

\begin{figure}[t]
\centering
\includegraphics[width=0.85\linewidth]{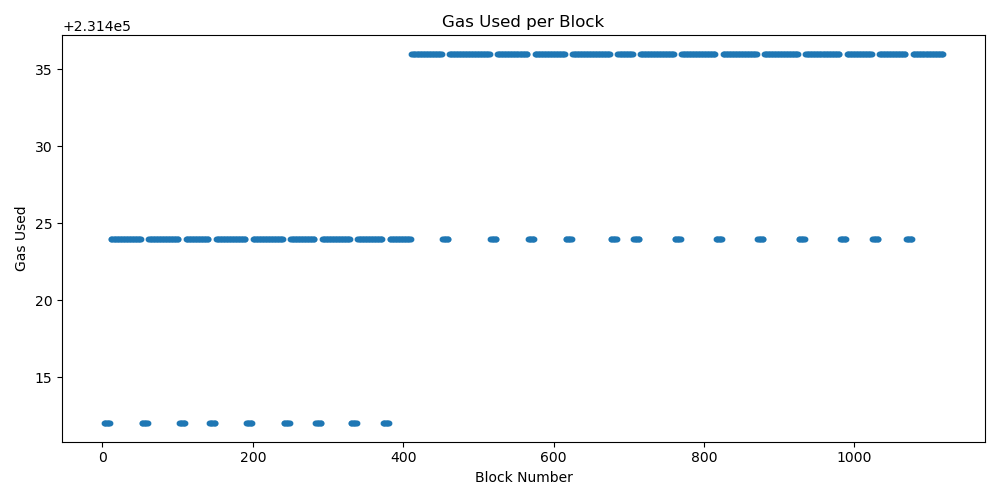}
\caption{Gas usage as a function of block number.}
\label{fig:gas-vs-block}
\end{figure}

Figures~\ref{fig:gas-usage-hist}--\ref{fig:gas-vs-block} illustrate the gas usage profile across all registration transactions.  
Because the contract executes a fixed sequence of operations per slide—computing a hash, assigning storage, and emitting a registration event—the gas consumed per call is effectively constant.

Across the full dataset, gas usage clusters tightly around $\approx 231{,}430$ gas per transaction.  
This constancy verifies that the contract is deterministic and that registration cost scales linearly with dataset size—a key property for long-term provenance tracking in educational repositories.

\subsubsection{Block Distribution and Registration Timeline}

\begin{figure}[t]
\centering
\includegraphics[width=0.8\linewidth]{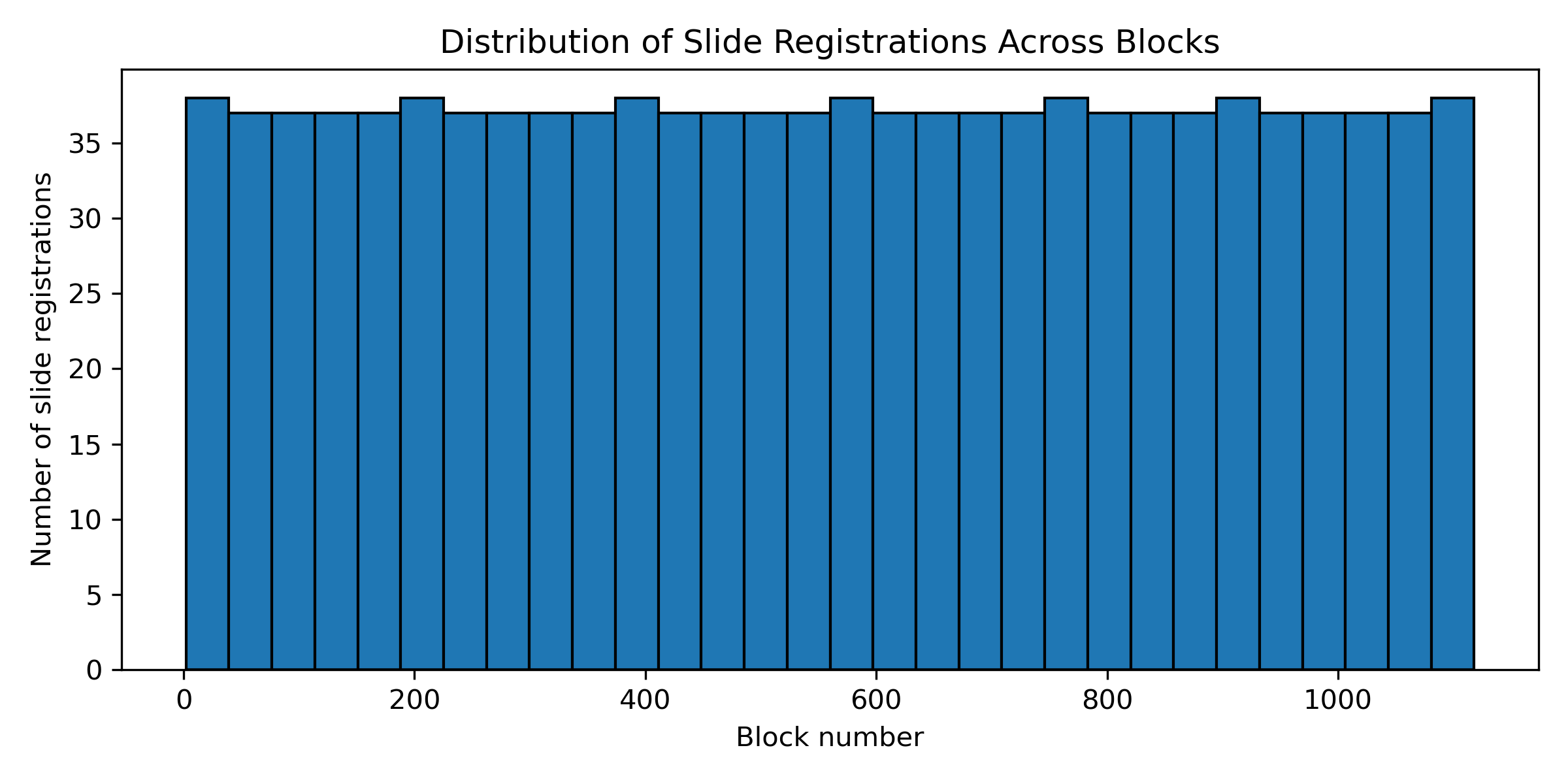}
\caption{Histogram of blocks containing provenance transactions.}
\label{fig:block-histogram}
\end{figure}

\begin{figure}[t]
\centering
\includegraphics[width=0.8\linewidth]{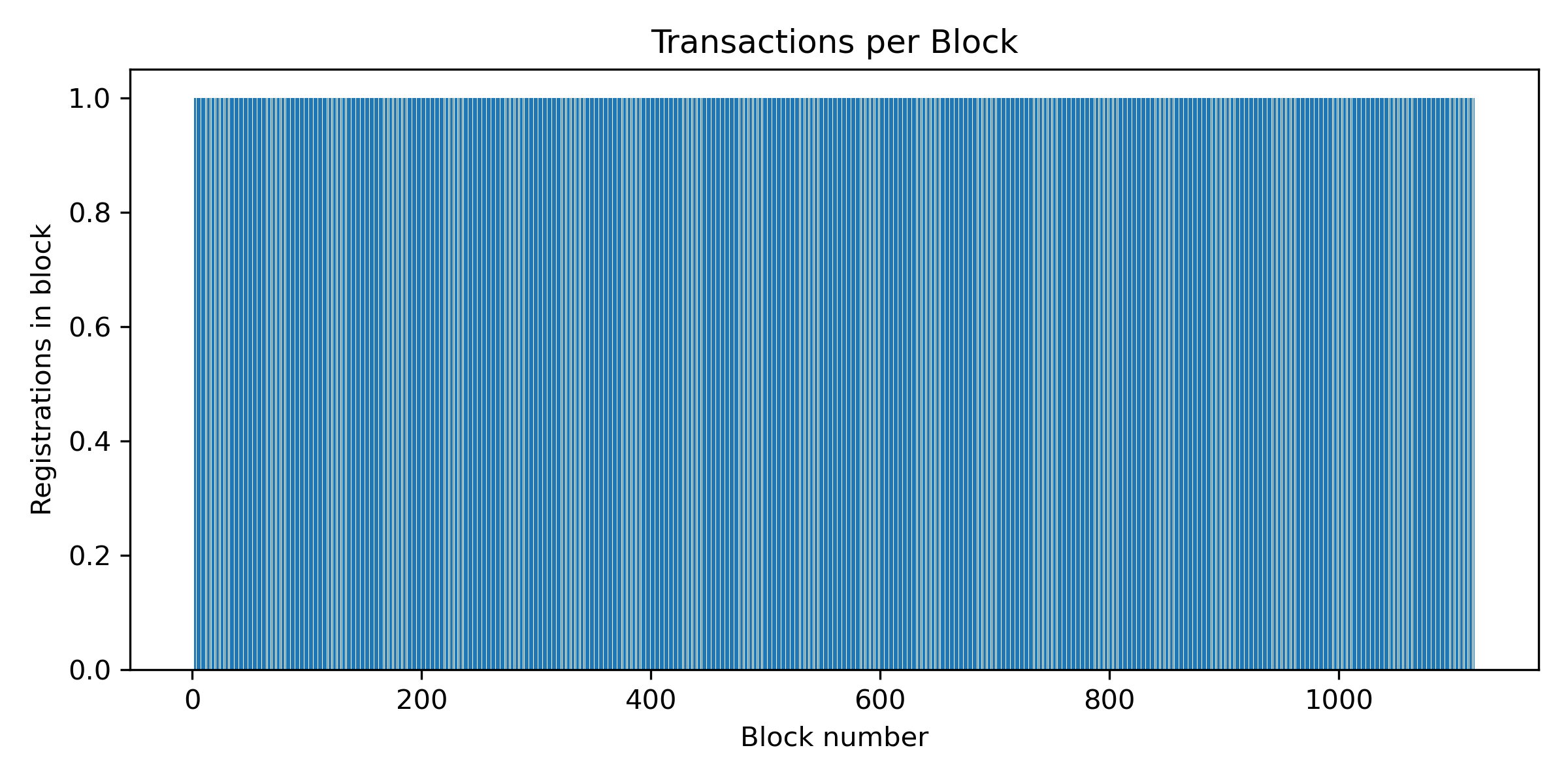}
\caption{Transactions per block during registration.}
\label{fig:tx-per-block}
\end{figure}

\begin{figure}[t]
\centering
\includegraphics[width=0.8\linewidth]{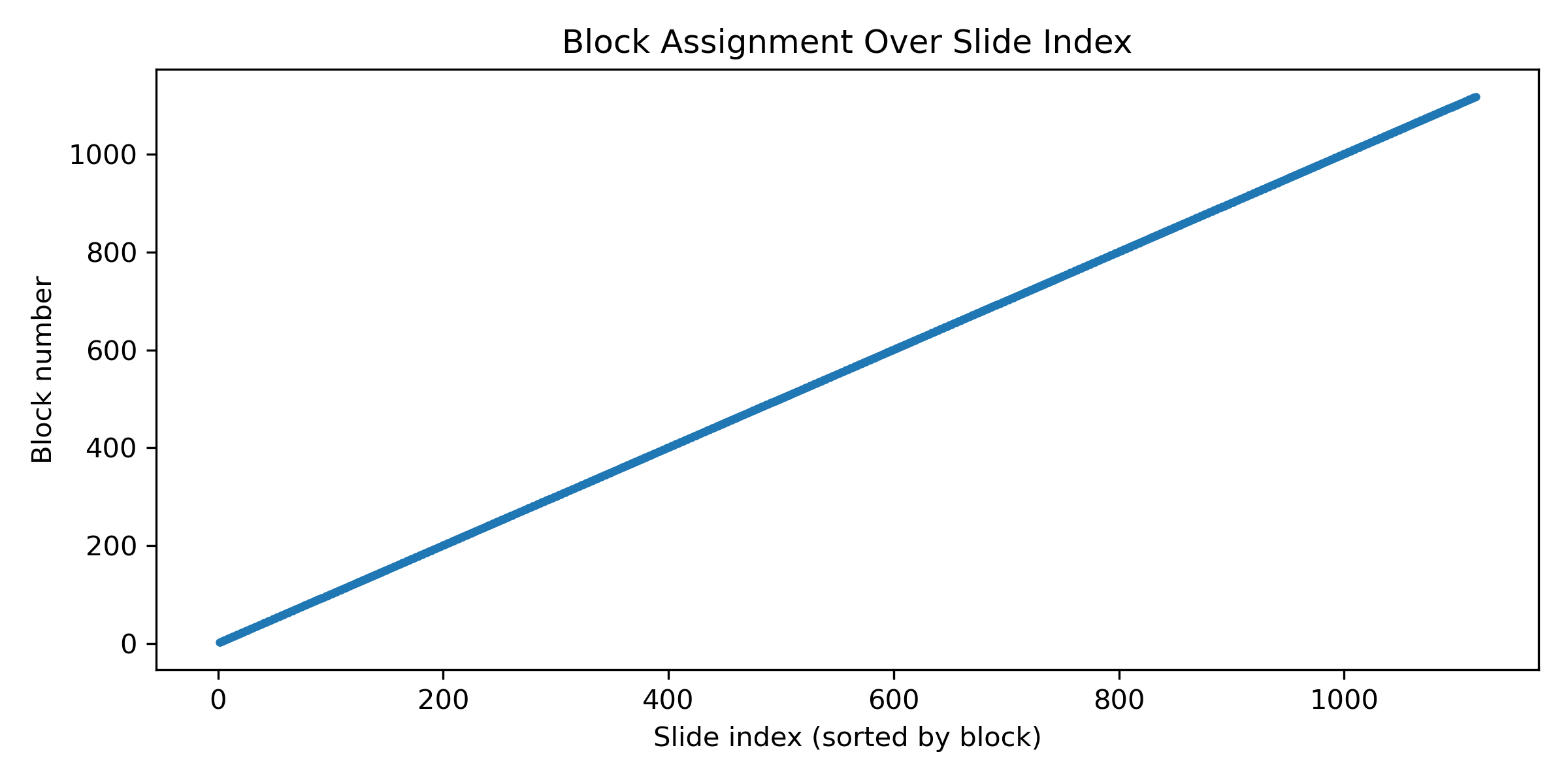}
\caption{Mapping between slide index and block height.}
\label{fig:block-vs-slide}
\end{figure}

\begin{figure}[t]
\centering
\includegraphics[width=0.85\linewidth]{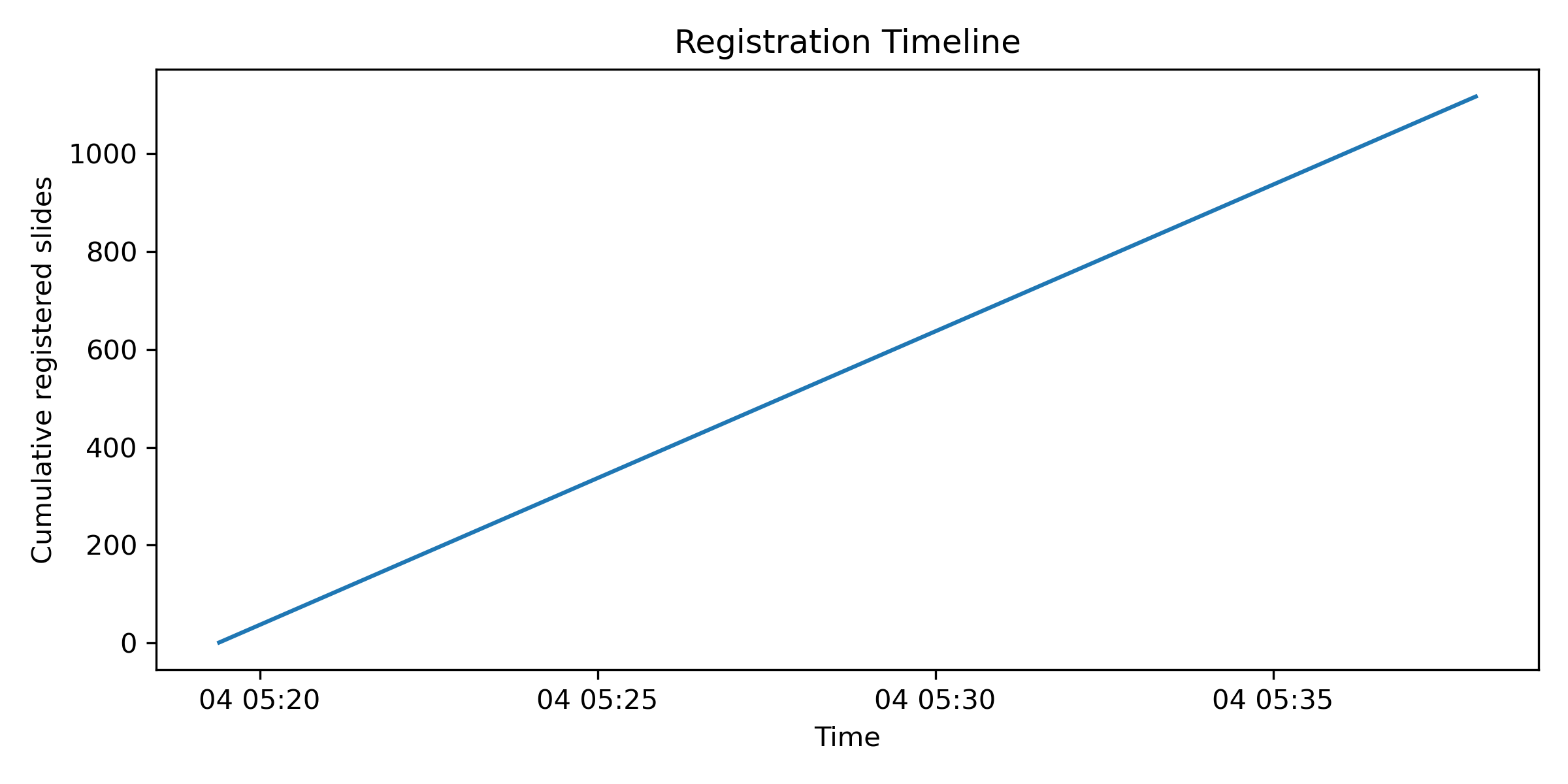}
\caption{Registration timestamp progression across the dataset.}
\label{fig:registration-timeline}
\end{figure}

We next examine the temporal and structural characteristics of how registrations were incorporated into the blockchain.  
Figures~\ref{fig:block-histogram}--\ref{fig:registration-timeline} report the block distribution, transactions per block, block--slide alignment, and timestamp progression.

Because the Hardhat miner seals a new block immediately upon receipt of a transaction, each slide maps to a unique block.  
This produces a strictly monotonic, near-perfect linear relationship between slide index and block height (Figure~\ref{fig:block-vs-slide}).  
The timestamp progression in Figure~\ref{fig:registration-timeline} further shows a stable throughput of approximately one slide per second.  
Such predictability simplifies deployment planning for larger-scale or real-time educational provenance systems.

\subsubsection{Cost Analysis}

\begin{figure}[t]
\centering
\includegraphics[width=0.8\linewidth]{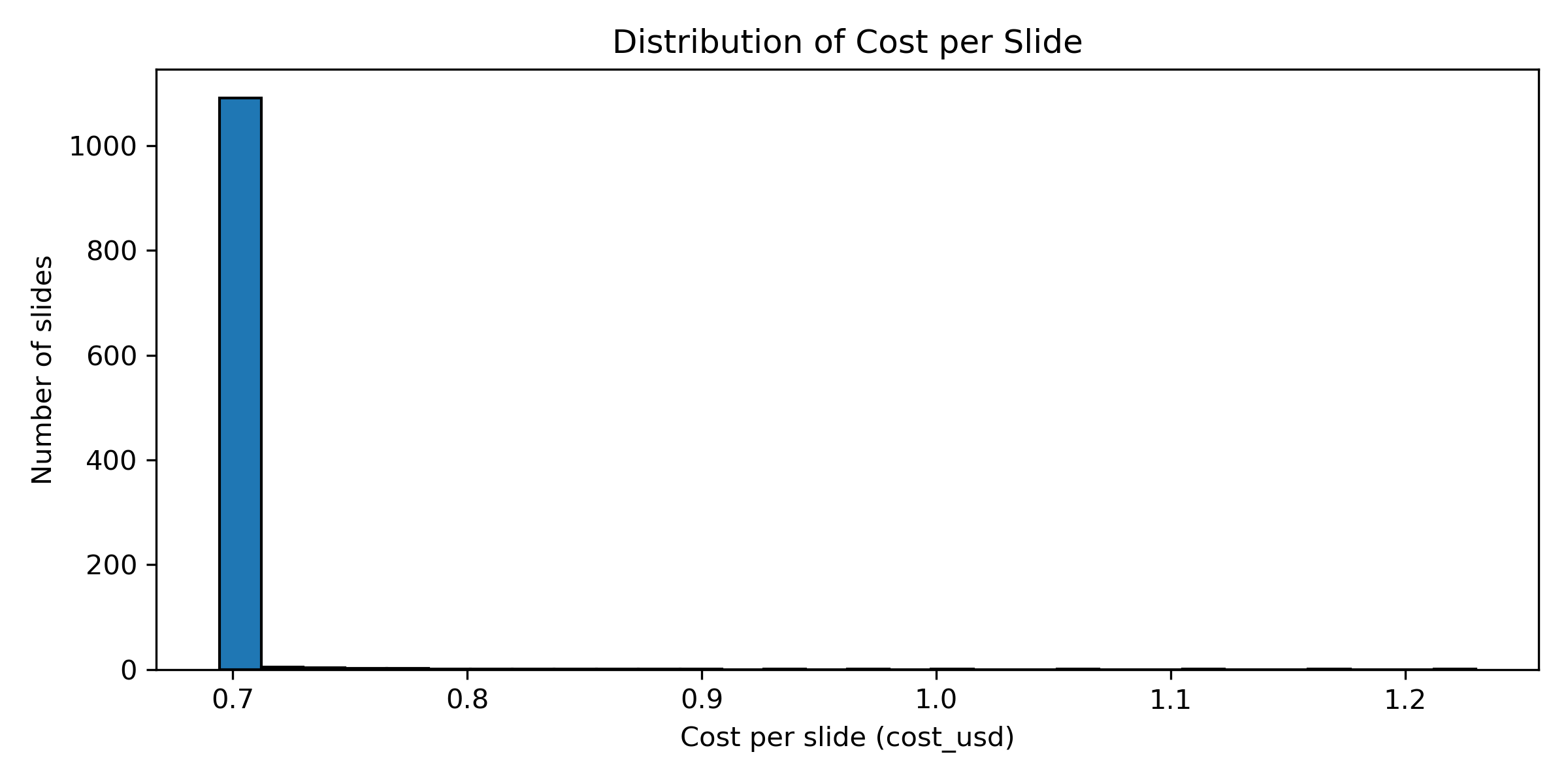}
\caption{Distribution of per-slide registration cost (USD).}
\label{fig:cost-histogram}
\end{figure}

\begin{figure}[t]
\centering
\includegraphics[width=0.8\linewidth]{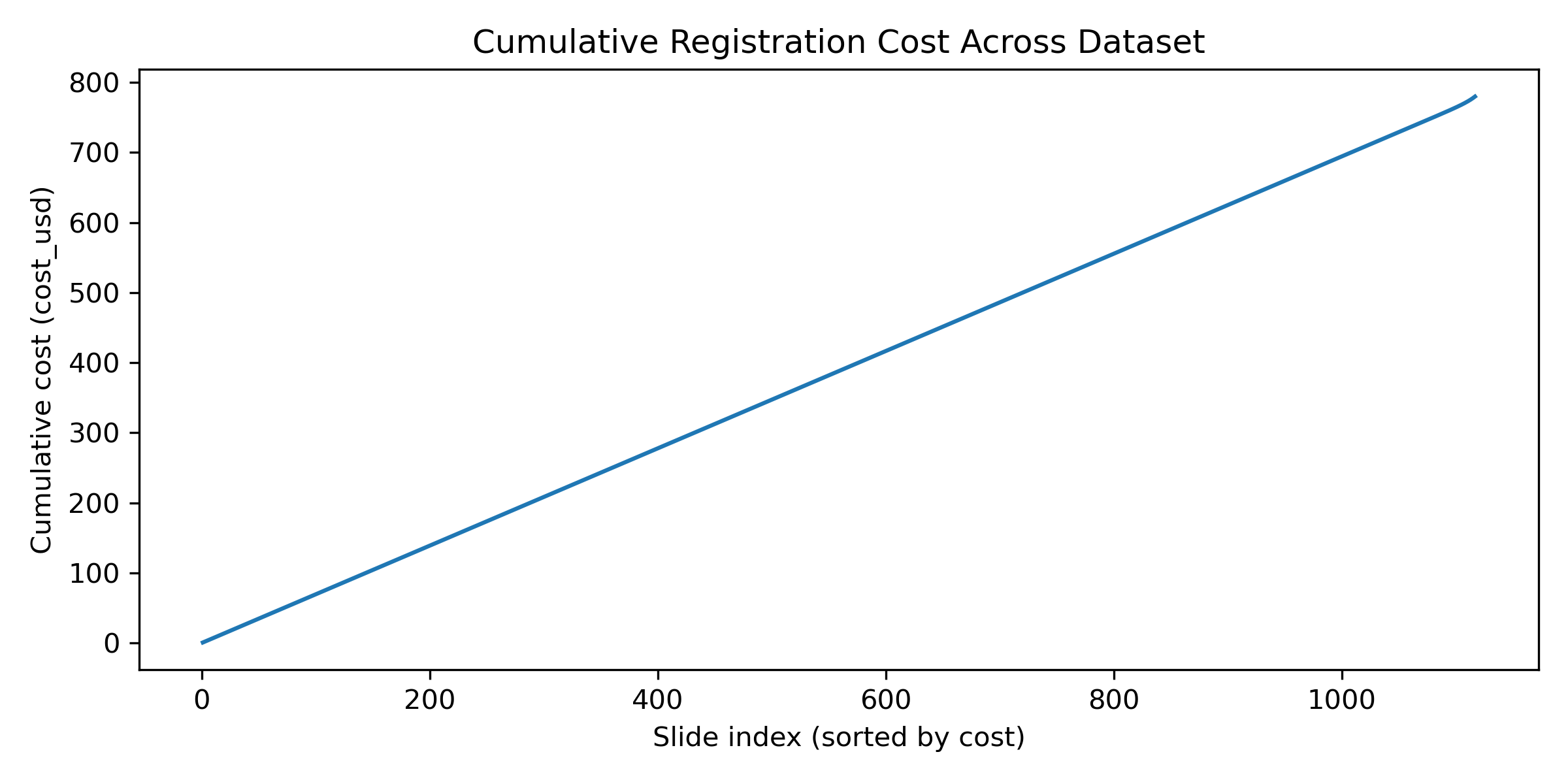}
\caption{Cumulative cost of registering the entire dataset.}
\label{fig:cumulative-cost}
\end{figure}

To assess financial feasibility, we convert gas usage to USD using a reference exchange rate of \$3000/ETH.  
Figures~\ref{fig:cost-histogram} and~\ref{fig:cumulative-cost} summarize the per-slide and cumulative registration costs.

On an Ethereum-like L1 environment, per-slide cost ranges from \$0.69 to \$1.23, totaling approximately \$780 for the 1,117-slide dataset.  
While L1 deployment remains relatively expensive, L2 ecosystems such as Optimism- or Polygon-like rollups reduce these costs by one to two orders of magnitude, making continuous provenance registration feasible for real-world educational pipelines.

\subsubsection{Scalability Simulation}

\begin{figure}[t]
\centering
\includegraphics[width=0.8\linewidth]{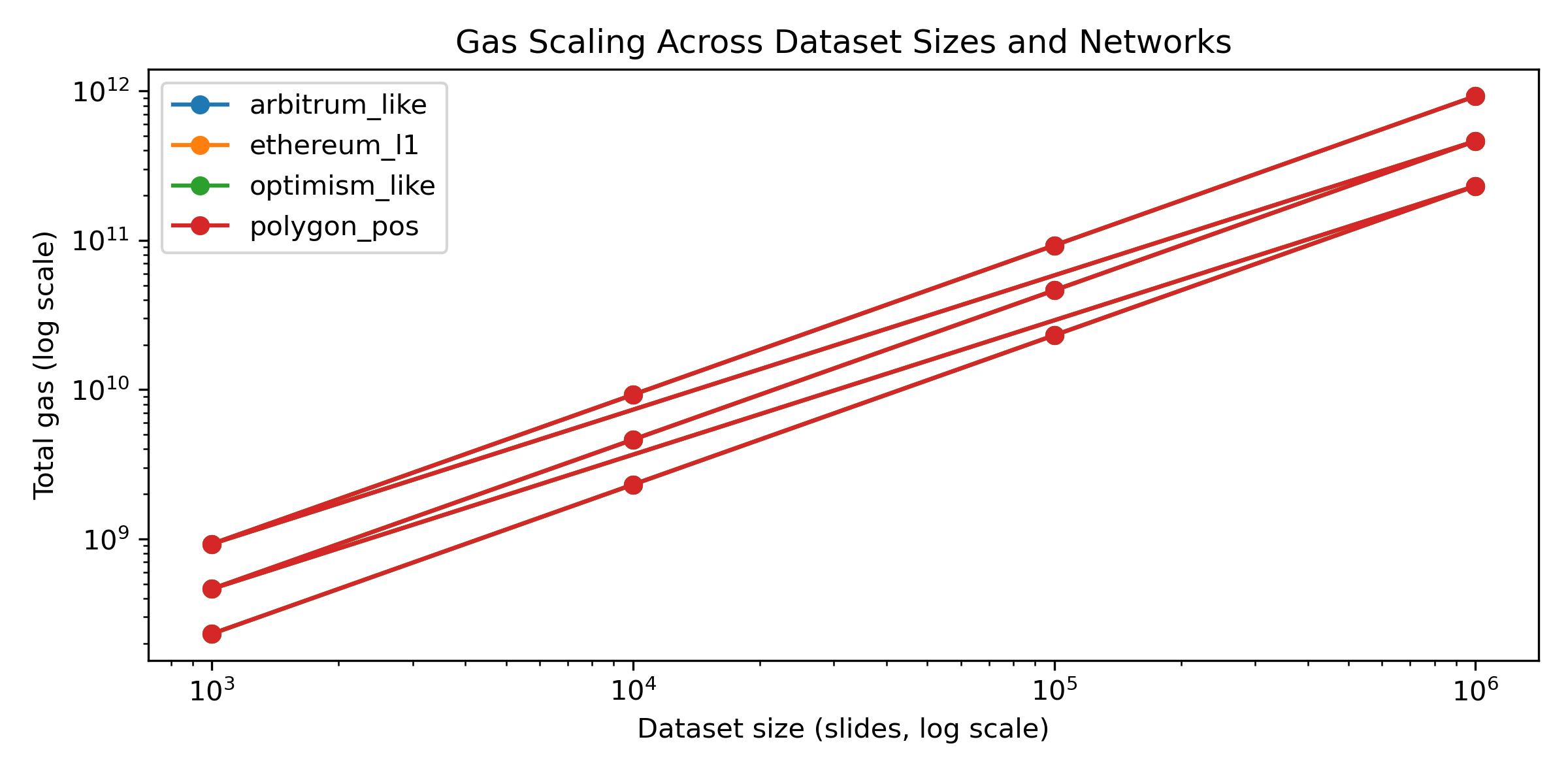}
\caption{Projected gas usage for datasets up to 1 million slides.}
\label{fig:scaling-gas}
\end{figure}

\begin{figure}[t]
\centering
\includegraphics[width=0.8\linewidth]{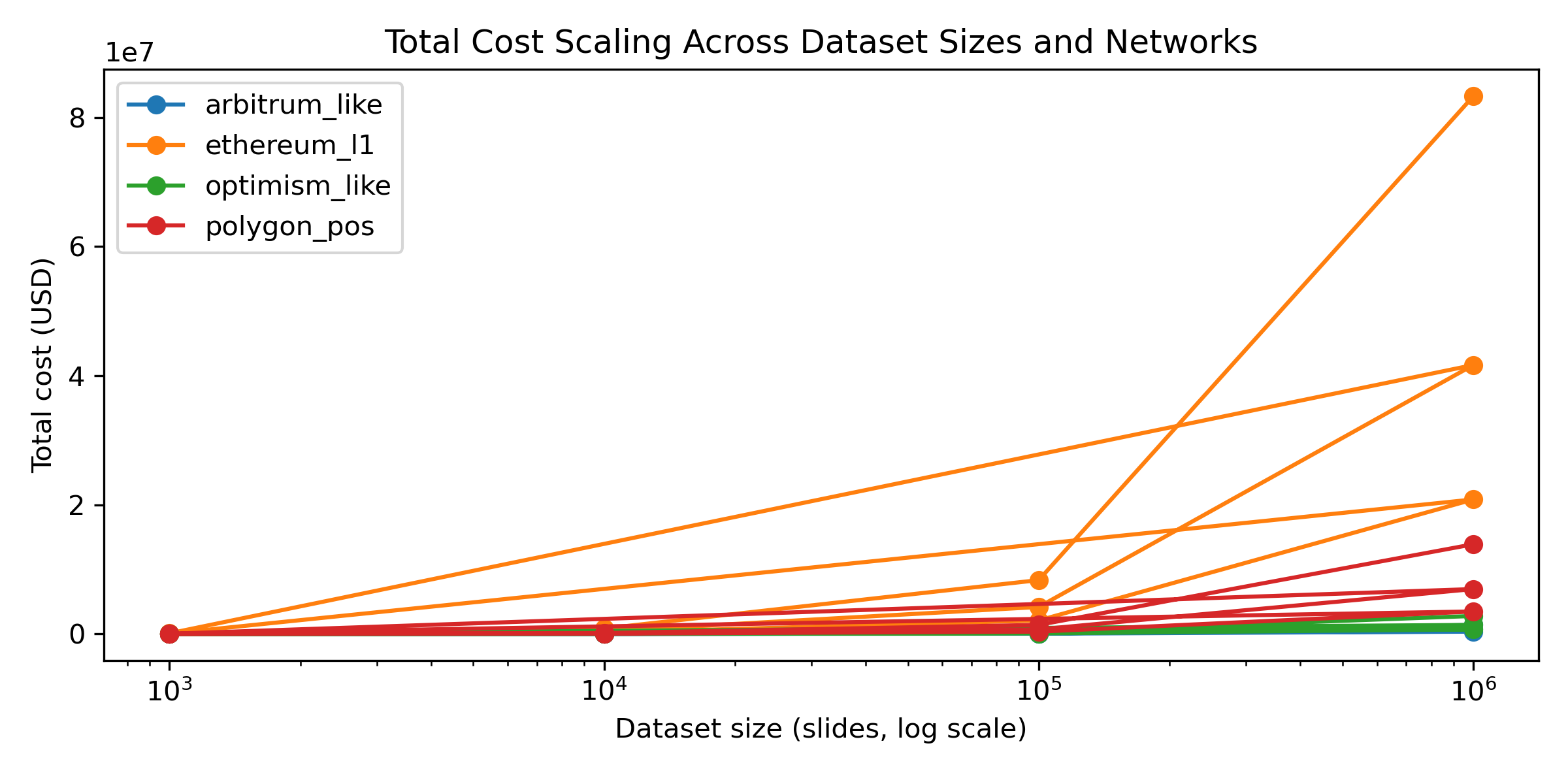}
\caption{Projected registration costs across networks.}
\label{fig:scaling-cost}
\end{figure}

To evaluate adoption, we extrapolate gas and financial costs for dataset sizes from $10^3$ to $10^6$ slides.  
Figures~\ref{fig:scaling-gas} and~\ref{fig:scaling-cost} present gas projections and corresponding multi-network cost curves.

Because the contract executes a constant number of storage operations per slide, total gas grows strictly linearly with dataset size.  
Under an L1 cost model, million-scale deployments would be costly, but on L2 networks the same operation remains economically practical.  
These results indicate that SlideChain can support course-level, curriculum-level, or MOOC-scale provenance infrastructures.
\subsubsection{Time-Gap Between File Creation and On-Chain Registration}

\begin{figure}[t]
\centering
\includegraphics[width=0.75\linewidth]{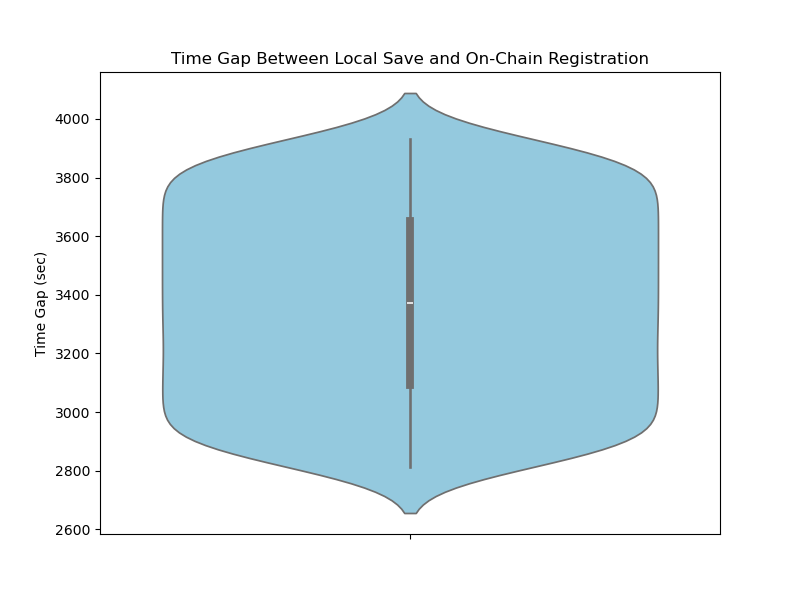}
\caption{Distribution of time gaps between local provenance generation and on-chain registration.}
\label{fig:timegap_violin}
\end{figure}

Figure~\ref{fig:timegap_violin} reports the distribution of time differences between local provenance-file creation times and their corresponding blockchain registration timestamps across all 1{,}117 slides.

The distribution is narrow and tightly concentrated, with time gaps centered around approximately 3{,}300 seconds and limited variance across the dataset.  
This behavior reflects stable batch registration of pre-generated provenance files under the deterministic Hardhat mining policy.

Because blockchain timestamps serve as canonical audit markers, such consistency ensures reliable chronological ordering and simplifies detection of delayed insertions, reordered records, or post hoc modifications during future verification.

\subsubsection{Tamper Detection Accuracy}

To evaluate robustness against malicious or accidental corruption, we applied controlled perturbations to a randomly selected subset of 20 provenance JSON files. Perturbations included modifying concept labels, altering or deleting relational triples, and injecting noise into evidence fields.

\begin{figure}[t]
\centering
\includegraphics[width=0.65\linewidth]{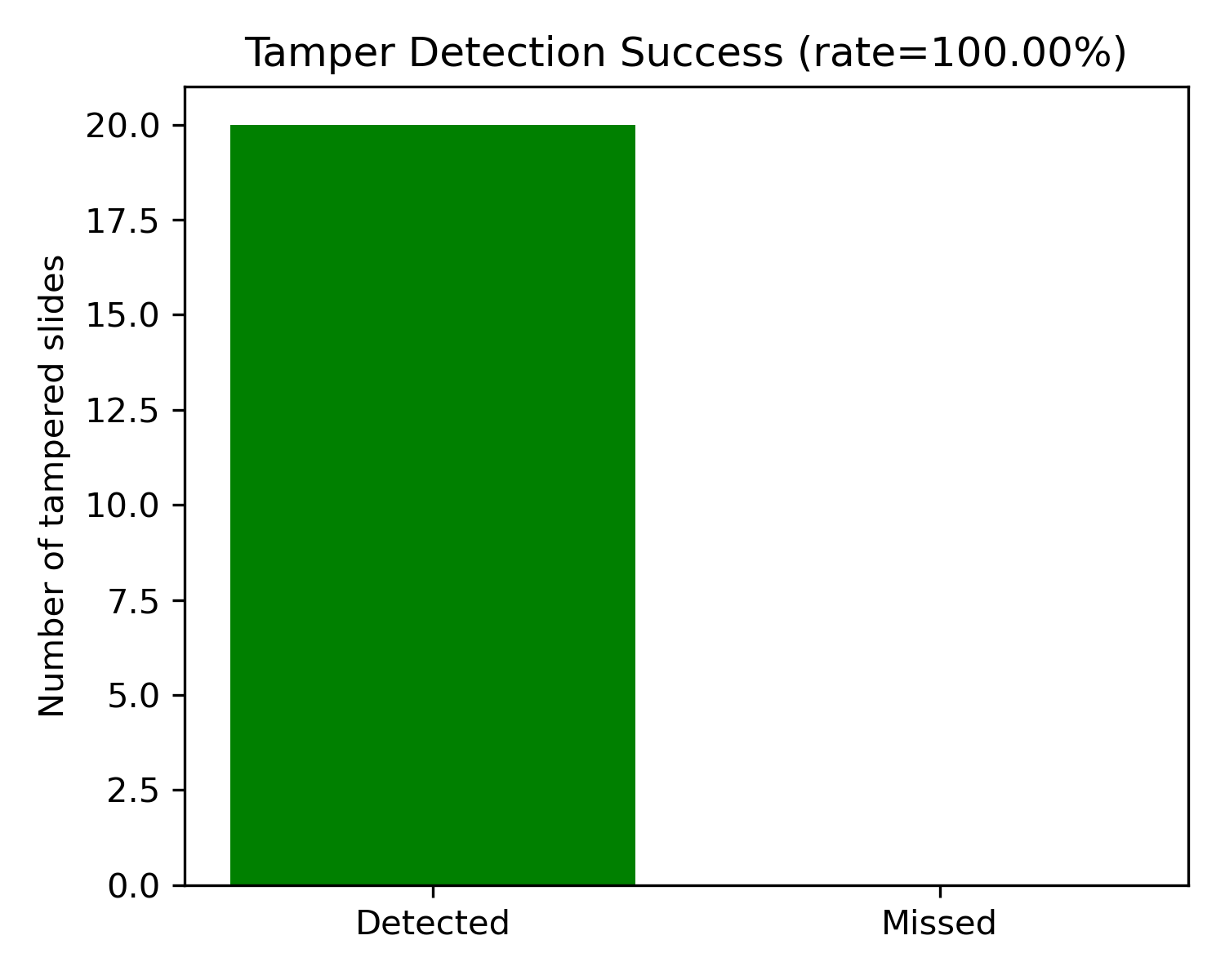}
\caption{Tamper detection performance under synthetic provenance corruption.}
\label{fig:tamper_detection_bar}
\end{figure}

As shown in Figure~\ref{fig:tamper_detection_bar}, all corrupted records were detected with 100\% accuracy.  
Because SlideChain relies on \texttt{keccak256} cryptographic digests, even a single-character modification produces a completely different hash, making undetected manipulation computationally infeasible at the dataset scale.

This experiment confirms that SlideChain provides strong, cryptographically grounded protection against provenance tampering.

\subsubsection{Reproducibility Across Two Independent Runs}

We assess semantic reproducibility by comparing two fully independent executions of the semantic extraction pipeline across all four VLMs and all 1{,}117 slides, yielding 4{,}468 model--slide comparisons.

Across all cases, both concept sets and relational triple sets matched exactly between runs, resulting in Jaccard similarities of 1.0 for every model and every slide.

Such perfect reproducibility is notable for multimodal VLM pipelines and ensures that any future semantic divergence can be cleanly attributed to model updates, preprocessing changes, or dataset modifications—rather than nondeterministic execution.  
When combined with immutable on-chain commitments, this property enables reliable long-term auditing of semantic drift.

\clearpage

\section{Discussion}
\label{sec:discussion}

This work presents SlideChain, a provenance infrastructure that couples multimodal semantic extraction with blockchain-backed verification to support auditability and reproducibility in educational AI systems. Rather than proposing improvements to semantic correctness itself, SlideChain focuses on making the \emph{outputs of semantic extraction verifiable, traceable, and comparable over time}. Across semantic analysis, blockchain evaluation, and reproducibility experiments, our results highlight both the opportunities and the limitations of current vision--language models when applied to scientific and engineering lecture material, and motivate provenance as a necessary system-level capability.

\subsection{Semantic Variability in Modern VLMs}

Although all four VLMs successfully extracted structured concepts and relational triples for the entire 1,117-slide dataset, their semantic outputs differed substantially. Disagreement distributions and Jaccard similarity matrices reveal only modest alignment at the concept level and, in many cases, near-zero overlap for relational triples across model pairs. Notably, this variability persists despite the slides being carefully curated, domain-specific, and pedagogically structured.

These findings suggest that semantic instability is not an edge case or artifact of noisy data, but a systemic property of current multimodal models, particularly when confronted with dense diagrams, mathematical expressions, imaging workflows, or visually complex scientific figures. For educational applications such as explanation generation, assessment support, or curriculum alignment, this instability implies that the output of a single VLM should not be treated as a definitive interpretation of instructional content.

SlideChain does not attempt to resolve or correct semantic disagreement. Instead, it makes such variability explicit, traceable, and auditable. By preserving model-specific outputs and anchoring them to immutable provenance records, the system enables structured comparison across models, runs, and time. In this framing, semantic divergence—often viewed as a failure mode—becomes an observable signal within a transparent and reproducible pipeline.

\noindent
Importantly, the value of semantic disagreement in SlideChain does not depend on establishing a single ground-truth interpretation of each slide. In educational AI systems, disagreement across models is itself a meaningful risk signal: it highlights regions of instructional content where automated interpretation is unstable, ambiguous, or sensitive to modeling choices. Such instability is particularly consequential in STEM education, where misleading or overconfident explanations can distort conceptual understanding. By making disagreement explicit and auditable rather than suppressing it, SlideChain enables provenance-aware, human-in-the-loop workflows in which instructors or domain experts can review, contextualize, and correct AI-generated semantics. In this sense, SlideChain is designed to support responsible deployment and oversight, not automated authority, aligning with emerging principles for trustworthy and transparent educational AI.

\subsection{Lecture-Level Patterns and Educational Implications}

Aggregated at the lecture level, semantic disagreement exhibits clear structure rather than uniform randomness. Slides involving complex imaging geometry, reconstruction equations, or visually dense multi-panel illustrations consistently show higher cross-model disagreement. These regions often correspond to conceptually challenging instructional material, indicating that semantic variability may itself carry pedagogical meaning.

This observation suggests two implications for educational AI systems:

\begin{itemize}
    \item \textbf{Instructor-facing analytics.}  
    Provenance-aware pipelines could flag slides with high semantic disagreement as candidates for enhanced explanation, additional annotation, or in-class emphasis.
    \item \textbf{Learner-facing adaptivity.}  
    For students, areas of high semantic instability may warrant uncertainty-aware explanations, multiple alternative interpretations, or interactive clarification rather than single, authoritative AI-generated responses.
\end{itemize}

By grounding these patterns in verifiable provenance records, SlideChain enables educational analytics that are not only informative, but auditable and reproducible.

\subsection{Blockchain Provenance as an Anchor for Semantic Auditability}

To clarify why SlideChain adopts a blockchain-backed provenance layer rather than conventional version control or logging mechanisms, Table~\ref{tab:provenance-comparison} contrasts these approaches along key integrity and trust dimensions. From a systems perspective, SlideChain demonstrates that blockchain-backed provenance can provide strong integrity guarantees with minimal overhead. The smart contract exhibits nearly constant gas usage per slide registration, confirming that the design remains lightweight even when applied to full-course datasets. Recording only the Keccak-256 hash on-chain strikes a deliberate balance: semantic records remain off-chain for flexibility and scalability, while the on-chain commitment serves as a permanent, tamper-evident reference.

Reproducibility experiments further emphasize this role. Under fixed prompts and execution environments, repeated extraction runs produce identical provenance records. Any future mismatch between a recomputed hash and the on-chain commitment therefore signals genuine semantic drift—whether due to model updates, preprocessing changes, or unintended modification. In this sense, SlideChain functions as a stability anchor, preserving a canonical semantic baseline against which future interpretations can be audited.

Such auditability is particularly important in educational settings, where lecture materials evolve, courses are revised, and AI models are continually updated. Provenance ensures that semantic interpretations remain reviewable, accountable, and historically traceable across these changes.

\begin{table}[t]
\centering
\caption{Comparison of provenance mechanisms for semantic auditability in multimodal AI systems.}
\label{tab:provenance-comparison}
\begin{tabular}{lccc}
\toprule
\textbf{Property} & \textbf{Git / DVC} & \textbf{Centralized Logging} & \textbf{Blockchain-backed (SlideChain)} \\
\midrule
Third-party verifiability & No & No & Yes \\
Tamper resistance & Limited$^{\dagger}$ & Limited$^{\dagger}$ & Strong (cryptographic) \\
Timestamp immutability & No & No & Yes \\
Trust assumptions & Repository owner & Service operator & Decentralized / verifiable \\
Silent overwrite detection & Partial & Partial & Guaranteed \\
Long-term auditability & Limited & Limited & Strong \\
\bottomrule
\end{tabular}
\\[0.5ex]
\footnotesize{$^{\dagger}$Requires trusting administrators and access controls rather than cryptographic guarantees.}
\end{table}

\subsection{Threat Model and Adversary Assumptions}
SlideChain is designed to provide integrity, auditability, and reproducibility guarantees for multimodal semantic records, rather than to ensure semantic correctness. Accordingly, our threat model focuses on adversaries who may attempt to modify, reorder, delete, or silently overwrite semantic provenance artifacts after extraction.

Specifically, SlideChain protects against post hoc tampering with off-chain semantic records, unauthorized modification of stored semantics, and ambiguity regarding when or under what conditions a semantic interpretation was produced. Any such modification is detectable through mismatch with the on-chain cryptographic commitment.

SlideChain does not protect against adversarial or erroneous model behavior at inference time, including hallucinated concepts, nonsensical outputs, or systematically biased interpretations generated by the underlying vision--language models. These risks are inherent to model design and data and must be addressed through model evaluation, human oversight, and domain expertise. In this sense, SlideChain provides a provenance and audit layer rather than a semantic validation or correctness mechanism.

\subsection{Trade-offs Between L1 and L2 Deployment}

Cost and scalability analyses highlight practical trade-offs in selecting a blockchain deployment environment. Ethereum-like L1 networks provide strong security and long-term archival guarantees, but at higher transaction costs, making them suitable for infrequent, high-assurance provenance checkpoints. In contrast, L2 rollups and sidechains—such as Optimism- or Polygon-style environments—reduce gas costs by one to two orders of magnitude, enabling frequent updates, registration, or near-real-time provenance logging.

These observations motivate a hybrid strategy: using L2 environments for routine provenance during active development, and periodically anchoring commitments to an L1 network for long-term archival assurance. This layered approach mirrors established practices in scientific data preservation and enterprise blockchain systems.
\subsection{Limitations}

Despite its strengths, SlideChain has several limitations that define its current scope and motivate future extensions:

\begin{itemize}
    \item \textbf{No semantic consensus or resolution.}  
    SlideChain quantifies and records semantic disagreement across VLMs but does not infer consensus semantics or correctness. Resolving disagreement would require principled uncertainty modeling, cross-model grounding, or ensemble reasoning, which remain open research challenges.

    \item \textbf{Dependence on pre-aligned multimodal inputs.}  
    The pipeline assumes that slide images and transcript segments are correctly aligned prior to semantic extraction. Misalignment could propagate errors into both semantic records and their provenance. Joint slide--text modeling or alignment verification mechanisms are not addressed in the current system.

    \item \textbf{Assumption of deterministic extraction behavior.}  
    Reproducibility results rely on empirically deterministic behavior under fixed prompts, environments, and execution settings. In deployments involving stochastic decoding, distributed inference, or heterogeneous hardware, exact reproducibility may not hold, although SlideChain would still detect semantic drift via hash mismatches.

    \item \textbf{Reliance on off-chain storage.}  
    For scalability and cost efficiency, full provenance records remain off-chain. Long-term accessibility therefore depends on the persistence and trustworthiness of external storage systems.

    \item \textbf{Simplified blockchain environment.}  
    All blockchain experiments were conducted on a deterministic, single-node Hardhat development chain. Real-world deployments must account for network latency, validator decentralization, gas-price volatility, and evolving EVM standards.

    \item \textbf{Limited domain scope.}  
    The empirical evaluation focuses on a single medical imaging course. Other STEM domains may exhibit different visual structures, semantic densities, or instability patterns, requiring additional validation to assess generality. At the same time, medical imaging lectures represent an unusually challenging testbed for multimodal semantic extraction. Slides in this domain are often diagram-heavy, symbol-rich, and densely structured, combining geometric illustrations, mathematical expressions, physical principles, and visual conventions that are difficult for current vision--language models to interpret reliably. In this sense, the chosen course serves as a stress test rather than an easy or narrowly constrained case, and the observed semantic variability is likely a lower bound on challenges that may arise in other technical STEM domains.

\end{itemize}

These limitations emphasize that SlideChain is intended as a foundational provenance layer for semantic auditability, rather than a semantic decision-making system or a comprehensive storage solution.

\subsection{Broader Impact and Significance}

As AI-generated interpretations are increasingly incorporated into STEM education, ensuring that such interpretations can be audited, reproduced, and trusted becomes a foundational requirement rather than an optional feature. SlideChain directly addresses this need by introducing verifiable provenance at the level of semantic outputs, enabling systematic tracking of how educational content is interpreted by evolving vision--language models.

By anchoring cryptographic commitments of slide-level semantics on-chain, SlideChain makes it possible to detect semantic drift, verify the integrity of extracted knowledge, and compare interpretations across models and over time. These capabilities support transparent auditing of AI-generated instructional content and provide a principled basis for understanding when, how, and why semantic interpretations change. Importantly, this infrastructure allows uncertainty-aware and provenance-guided educational tools to be built without assuming that any single model provides a definitive or stable interpretation.

More broadly, this work argues that provenance should be treated as a \emph{first-class system component} in educational AI. Rather than tying trust to specific models or architectures, SlideChain anchors meaning itself to a tamper-proof ledger, providing accountability and reproducibility independent of future advances in model accuracy. As AI systems play an increasingly central role in teaching, assessment, and curriculum design, such provenance-aware foundations will be essential for responsible, transparent, and trustworthy educational deployment.

\subsection{Ethical Considerations and Responsible Use}

SlideChain is designed to provide verifiable integrity and auditability of semantic outputs, not to assess their correctness, pedagogical quality, or clinical validity. A potential misuse of provenance systems is treating immutability as an implicit endorsement of semantic accuracy, thereby lending undue authority to flawed or misleading AI-generated interpretations.

SlideChain explicitly avoids this pitfall by design. The system records and exposes semantic variability, disagreement, and drift across models rather than collapsing them into a single authoritative representation. Provenance is therefore used to support transparency and accountability, not automated trust. By making semantic uncertainty visible and auditable, SlideChain encourages human oversight and responsible use in educational settings, where expert judgment remains essential.

\section{Conclusion}
\label{sec:conclusion}

This paper introduced \textbf{SlideChain}, designed to ensure the integrity, auditability, and long-term reproducibility of multimodal semantic extraction pipelines for educational content. Using the SlideChain Slides Dataset of 1,117 medical imaging lecture slides and four state-of-the-art vision--language models, we presented, to the best of our knowledge, the first empirical study of semantic variability in real-world STEM lecture material. Our results reveal substantial cross-model divergence in both concept extraction and relational grounding, underscoring that modern VLMs—despite their impressive generative capabilities—do not yet provide stable or authoritative interpretations of scientific slides.
SlideChain addresses this challenge by anchoring cryptographic commitments of slide-level semantic records to an Ethereum-compatible blockchain. While simple hashing or local logging can detect accidental file changes, they cannot provide immutable timestamps, third-party verifiability, or protection against silent overwriting. In contrast, blockchain-backed commitments establish an append-only, tamper-evident ledger that decouples trust from any single system owner and enables independent auditing of semantic interpretations over time. By storing only compact Keccak-256 hashes on-chain and retaining full semantic records off-chain, SlideChain achieves this integrity guarantee without sacrificing flexibility or scalability.
Our evaluations demonstrate that the system exhibits predictable, constant gas usage, near-linear scalability, and stable throughput of approximately one slide per second under batch registration. Tamper-detection experiments achieved 100\% accuracy, and dual-run reproducibility tests produced perfect Jaccard similarity across all 4,468 model--slide combinations, confirming deterministic behavior under controlled execution settings.
Taken together, these results establish SlideChain as a practical and extensible provenance infrastructure for multimodal educational AI. Rather than attempting to resolve semantic correctness, SlideChain provides a verifiable semantic baseline against which model behavior, drift, and disagreement can be transparently audited. As AI-generated interpretations become increasingly embedded in STEM pedagogy and digital learning ecosystems, such provenance guarantees will be essential for maintaining trust, accountability, and long-term reproducibility.

\section{Future Work}
\label{sec:future-works}

Several extensions can build directly on the current limitations of SlideChain and broaden its applicability as a general-purpose provenance infrastructure.

\begin{itemize}[leftmargin=12pt]
    \item \textbf{Merkle-based aggregation for scalable provenance.}  
    To improve scalability and batch verification, future versions may aggregate slide-level commitments into Merkle trees at the lecture or course level. This would enable compact audit proofs, efficient verification of large collections, and structured checkpointing aligned with educational units.

    \item \textbf{Decentralized storage for long-term accessibility.}  
    To reduce reliance on local or centralized storage, SlideChain can be extended to integrate decentralized storage systems such as IPFS, Filecoin, or Arweave. Coupled with on-chain commitments, this would provide global availability, redundancy, and long-term persistence of provenance records.

    \item \textbf{Cross-model consensus and stability metrics.}  
    The semantic disagreement observed across VLMs motivates the development of consensus-based scoring, stability metrics, or uncertainty-aware aggregation. Such tools could help identify conceptually unstable slide regions, guide targeted instructor review, or support ensemble-based semantic interpretation without assuming a single authoritative model.

    \item \textbf{Versioned and incremental provenance tracking.}  
    As models, prompts, and preprocessing pipelines evolve, semantic interpretations will drift. Supporting append-only version chains, per-model lineage tracking, and structured semantic diffs would enable systematic study of how model updates influence content understanding over time.

    \item \textbf{Automated model-drift monitoring.}  
    Because SlideChain establishes immutable semantic baselines, future systems can automatically detect model drift by comparing new extractions against registered commitments. This capability is broadly applicable to educational and safety-critical AI pipelines that require stable, auditable behavior.

    \item \textbf{User-facing provenance interfaces.}  
    Interactive dashboards exposing semantic variability, model agreement, and provenance status to instructors and learners would enable transparent inspection of AI-generated interpretations and support trust-aware educational workflows.
\end{itemize}

Together, these directions position SlideChain as a foundation for scalable, transparent, and reproducible multimodal AI infrastructure, extensible beyond education to scientific, clinical, and enterprise settings where verifiable semantic interpretation is essential.

\section*{Acknowledgment}
We thank Clara Reyes for helpful comments and suggestions that improved the clarity and presentation of the manuscript.

\section*{Funding Statement}
This research received no external funding.


\printbibliography

@article{du2022survey,
  title={A survey of vision-language pre-trained models},
  author={Du, Yifan and Liu, Zikang and Li, Junyi and Zhao, Wayne Xin},
  journal={arXiv preprint arXiv:2202.10936},
  year={2022}
}

@article{liu2024survey,
  title={A survey on hallucination in large vision-language models},
  author={Liu, Hanchao and Xue, Wenyuan and Chen, Yifei and Chen, Dapeng and Zhao, Xiutian and Wang, Ke and Hou, Liping and Li, Rongjun and Peng, Wei},
  journal={arXiv preprint arXiv:2402.00253},
  year={2024}
}

@article{sahoo2024comprehensive,
  title={A comprehensive survey of hallucination in large language, image, video and audio foundation models},
  author={Sahoo, Pranab and Meharia, Prabhash and Ghosh, Akash and Saha, Sriparna and Jain, Vinija and Chadha, Aman},
  journal={arXiv preprint arXiv:2405.09589},
  year={2024}
}

@article{semmelrock2023reproducibility,
  title={Reproducibility in machine learning-driven research},
  author={Semmelrock, Harald and Kopeinik, Simone and Theiler, Dieter and Ross-Hellauer, Tony and Kowald, Dominik},
  journal={arXiv preprint arXiv:2307.10320},
  year={2023}
}

@article{beam2020challenges,
  title={Challenges to the reproducibility of machine learning models in health care},
  author={Beam, Andrew L and Manrai, Arjun K and Ghassemi, Marzyeh},
  journal={Jama},
  volume={323},
  number={4},
  pages={305--306},
  year={2020},
  publisher={American Medical Association}
}

@misc{liu2023visualinstructiontuning,
      title={Visual Instruction Tuning}, 
      author={Haotian Liu and Chunyuan Li and Qingyang Wu and Yong Jae Lee},
      year={2023},
      eprint={2304.08485},
      archivePrefix={arXiv},
      primaryClass={cs.CV},
      url={https://arxiv.org/abs/2304.08485}, 
}

@misc{bai2023qwenvlversatilevisionlanguagemodel,
      title={Qwen-VL: A Versatile Vision-Language Model for Understanding, Localization, Text Reading, and Beyond}, 
      author={Jinze Bai and Shuai Bai and Shusheng Yang and Shijie Wang and Sinan Tan and Peng Wang and Junyang Lin and Chang Zhou and Jingren Zhou},
      year={2023},
      eprint={2308.12966},
      archivePrefix={arXiv},
      primaryClass={cs.CV},
      url={https://arxiv.org/abs/2308.12966}, 
}

@misc{zhu2025internvl3exploringadvancedtraining,
      title={InternVL3: Exploring Advanced Training and Test-Time Recipes for Open-Source Multimodal Models}, 
      author={Jinguo Zhu and Weiyun Wang and Zhe Chen and Zhaoyang Liu and Shenglong Ye and Lixin Gu and Hao Tian and Yuchen Duan and Weijie Su and Jie Shao and Zhangwei Gao and Erfei Cui and Xuehui Wang and Yue Cao and Yangzhou Liu and Xingguang Wei and Hongjie Zhang and Haomin Wang and Weiye Xu and Hao Li and Jiahao Wang and Nianchen Deng and Songze Li and Yinan He and Tan Jiang and Jiapeng Luo and Yi Wang and Conghui He and Botian Shi and Xingcheng Zhang and Wenqi Shao and Junjun He and Yingtong Xiong and Wenwen Qu and Peng Sun and Penglong Jiao and Han Lv and Lijun Wu and Kaipeng Zhang and Huipeng Deng and Jiaye Ge and Kai Chen and Limin Wang and Min Dou and Lewei Lu and Xizhou Zhu and Tong Lu and Dahua Lin and Yu Qiao and Jifeng Dai and Wenhai Wang},
      year={2025},
      eprint={2504.10479},
      archivePrefix={arXiv},
      primaryClass={cs.CV},
      url={https://arxiv.org/abs/2504.10479}, 
}

@misc{tanaka2023slidevqadatasetdocumentvisual,
      title={SlideVQA: A Dataset for Document Visual Question Answering on Multiple Images}, 
      author={Ryota Tanaka and Kyosuke Nishida and Kosuke Nishida and Taku Hasegawa and Itsumi Saito and Kuniko Saito},
      year={2023},
      eprint={2301.04883},
      archivePrefix={arXiv},
      primaryClass={cs.CL},
      url={https://arxiv.org/abs/2301.04883}, 
}

@inproceedings{lee2023lecture,
  title={Lecture presentations multimodal dataset: Towards understanding multimodality in educational videos},
  author={Lee, Dong Won and Ahuja, Chaitanya and Liang, Paul Pu and Natu, Sanika and Morency, Louis-Philippe},
  booktitle={Proceedings of the IEEE/CVF International Conference on Computer Vision},
  pages={20087--20098},
  year={2023}
}

@inproceedings{zhang2025towards,
  title={Towards Comprehensive Lecture Slides Understanding: Large-scale Dataset and Effective Method},
  author={Zhang, Enming and Li, Yuzhe and Liu, Yuliang and Zhu, Yingying and Bai, Xiang},
  booktitle={Proceedings of the IEEE/CVF International Conference on Computer Vision},
  pages={4455--4464},
  year={2025}
}

@misc{liu2024surveyhallucinationlargevisionlanguage,
      title={A Survey on Hallucination in Large Vision-Language Models}, 
      author={Hanchao Liu and Wenyuan Xue and Yifei Chen and Dapeng Chen and Xiutian Zhao and Ke Wang and Liping Hou and Rongjun Li and Wei Peng},
      year={2024},
      eprint={2402.00253},
      archivePrefix={arXiv},
      primaryClass={cs.CV},
      url={https://arxiv.org/abs/2402.00253}, 
}

@inproceedings{sahoo-etal-2024-comprehensive,
    title = "A Comprehensive Survey of Hallucination in Large Language, Image, Video and Audio Foundation Models",
    author = "Sahoo, Pranab  and
      Meharia, Prabhash  and
      Ghosh, Akash  and
      Saha, Sriparna  and
      Jain, Vinija  and
      Chadha, Aman",
    editor = "Al-Onaizan, Yaser  and
      Bansal, Mohit  and
      Chen, Yun-Nung",
    booktitle = "Findings of the Association for Computational Linguistics: EMNLP 2024",
    month = nov,
    year = "2024",
    address = "Miami, Florida, USA",
    publisher = "Association for Computational Linguistics",
    url = "https://aclanthology.org/2024.findings-emnlp.685/",
    doi = "10.18653/v1/2024.findings-emnlp.685",
    pages = "11709--11724"
}

@article{belhajjame2013prov,
  title={Prov-dm: The prov data model},
  author={Belhajjame, Khalid and B’Far, Reza and Cheney, James and Coppens, Sam and Cresswell, Stephen and Gil, Yolanda and Groth, Paul and Klyne, Graham and Lebo, Timothy and McCusker, Jim and others},
  journal={W3C Recommendation},
  volume={14},
  pages={15--16},
  year={2013},
  publisher={World Wide Web Consortium (W3C)}
}

@misc{semmelrock2023reproducibilitymachinelearningdrivenresearch,
      title={Reproducibility in Machine Learning-Driven Research}, 
      author={Harald Semmelrock and Simone Kopeinik and Dieter Theiler and Tony Ross-Hellauer and Dominik Kowald},
      year={2023},
      eprint={2307.10320},
      archivePrefix={arXiv},
      primaryClass={cs.LG},
      url={https://arxiv.org/abs/2307.10320}, 
}

@misc{kapoor2022leakagereproducibilitycrisismlbased,
      title={Leakage and the Reproducibility Crisis in ML-based Science}, 
      author={Sayash Kapoor and Arvind Narayanan},
      year={2022},
      eprint={2207.07048},
      archivePrefix={arXiv},
      primaryClass={cs.LG},
      url={https://arxiv.org/abs/2207.07048}, 
}

@misc{Tuychiev2024DVC,
  author       = {Bex Tuychiev},
  title        = {The Complete Guide to Data Version Control With DVC},
  howpublished = {\url{https://www.datacamp.com/tutorial/data-version-control-dvc}},
  year         = {2024},
  month        = {Jul 14},
  note         = {Accessed: 2025-12-08}
}

@incollection{machine-learning-in-production-ch24,
  title        = {Versioning, Provenance, and Reproducibility},
  booktitle    = {Machine Learning in Production: From Models to Products},
  author       = {Christian Kästner},
  year         = {2025},
  publisher    = {MIT Press},
  url          = {https://mlip-cmu.github.io/book/24-versioning-provenance-and-reproducibility.html},
  note         = {Chapter 24, accessed 2025-12-08}
}

@article{todd2016opentimestamps,
  title={Opentimestamps: Scalable, trust-minimized, distributed timestamping with bitcoin},
  author={Todd, P},
  journal={Peter Todd [Internet]},
  volume={15},
  year={2016}
}

@inproceedings{liang2017provchain,
  title={Provchain: A blockchain-based data provenance architecture in cloud environment with enhanced privacy and availability},
  author={Liang, Xueping and Shetty, Sachin and Tosh, Deepak and Kamhoua, Charles and Kwiat, Kevin and Njilla, Laurent},
  booktitle={2017 17th IEEE/ACM International Symposium on Cluster, Cloud and Grid Computing (CCGRID)},
  pages={468--477},
  year={2017},
  organization={IEEE}
}

@article{ruan2019fine,
  title={Fine-grained, secure and efficient data provenance on blockchain systems},
  author={Ruan, Pingcheng and Chen, Gang and Dinh, Tien Tuan Anh and Lin, Qian and Ooi, Beng Chin and Zhang, Meihui},
  journal={Proceedings of the VLDB Endowment},
  volume={12},
  number={9},
  pages={975--988},
  year={2019},
  publisher={VLDB Endowment}
}

@article{sun2022bstprov,
  title={BSTProv: blockchain-based secure and trustworthy data provenance sharing},
  author={Sun, Lian-Shan and Bai, Xue and Zhang, Chao and Li, Yang and Zhang, Yong-Bin and Guo, Wen-Qiang},
  journal={Electronics},
  volume={11},
  number={9},
  pages={1489},
  year={2022},
  publisher={MDPI}
}

@inproceedings{al2021scichain,
  title={Scichain: Blockchain-enabled lightweight and efficient data provenance for reproducible scientific computing},
  author={Al-Mamun, Abdullah and Yan, Feng and Zhao, Dongfang},
  booktitle={2021 IEEE 37th international conference on data engineering (ICDE)},
  pages={1853--1858},
  year={2021},
  organization={IEEE}
}

@article{akbarfam2024sok,
  title={SOK: Blockchain for Provenance},
  author={Akbarfam, Asma Jodeiri and Maleki, Hoda},
  journal={arXiv preprint arXiv:2407.17699},
  year={2024}
}

\clearpage

\appendix
\section*{Appendix}

This appendix provides reproducibility resources and additional implementation details supporting the main paper. Only essential supplementary content is included to maintain clarity and avoid unnecessary length.




\section{Reproducibility Resources}

To support full replication of all SlideChain experiments, we provide the complete codebase, provenance artifacts, blockchain configurations, and analysis outputs used in this paper. These materials enable end-to-end reproduction of semantic extraction, provenance hashing, smart-contract registration, and all statistical evaluations.

The shared resources include:

\begin{itemize}
    \item the SlideChain smart contract and Hardhat deployment environment,
    \item Python scripts for semantic extraction, normalization, provenance generation, and analysis,
    \item all CSV files used in the paper (semantic disagreement statistics, Jaccard matrices, blockchain metrics, integrity and scalability logs),
    \item all figures appearing in the main text and appendix,
    \item slide-level provenance JSON files for all 1,117 slides, organized by lecture,
    \item complete environment configuration files for Python and Hardhat.
\end{itemize}

\paragraph{Repository Location.}  
A public GitHub repository containing all project materials can be found here: \url{https://github.com/manikm-114/SlideChain}

\paragraph{Top-Level Layout.}
A high-level view of the reproducibility package is shown below:

\begin{verbatim}
SlideChain/
 |-- SlideChain/      # Core blockchain + analysis project
 |-- Lectures/        # Slide-level provenance JSON files
\end{verbatim}

\paragraph{SlideChain Project Directory.}
The \texttt{SlideChain/SlideChain} directory contains all components required to
redeploy the smart contract, rerun blockchain registration, and regenerate the figures
and tables reported in the manuscript:

\begin{verbatim}
SlideChain/SlideChain/
 |-- contracts/           # SlideChain.sol smart contract
 |-- scripts/             # Deployment + registration utilities
 |-- analysis_results/    # CSV outputs from semantic + chain analysis
 |-- chain_results/       # Gas, cost, throughput, and integrity logs
 |-- figures/             # All provenance + blockchain figures
 |-- environment/         # requirements.txt, hardhat.config.js
 |-- (standard Hardhat build files omitted for brevity)
\end{verbatim}

Only essential directories are shown; autogenerated caches, ABI metadata, and
other Hardhat boilerplate are intentionally omitted for clarity.

\paragraph{Lecture-Level Provenance Directory.}
The slide-level semantic provenance records used throughout the paper are stored in:

\begin{verbatim}
SlideChain/Lectures/
 |-- by_slide/
      |-- Lecture 1/
      |    |-- Slide1.json
      |    |-- Slide2.json
      |    |-- ...
      |-- Lecture 2/
      |    |-- ...
      |-- ...
      |-- Lecture 23/
\end{verbatim}

Each \texttt{SlideX.json} file contains the unified provenance record for a single slide
(concepts, triples, evidence, metadata for all four VLMs).  
These are the exact artifacts whose keccak256 hashes are anchored on-chain by the SlideChain contract.

\paragraph{Reproducibility Checklist.}
We follow standard reproducibility practices for machine learning and systems research:

\begin{itemize}
    \item Deterministic preprocessing and fixed random seeds are used for all experiments.
    \item All VLM prompts, inference parameters, and model identifiers are included in the repository.
    \item Both extraction runs used in the reproducibility analysis are preserved in versioned directories.
    \item Blockchain registration can be fully regenerated on any EVM-compatible local chain (e.g., Hardhat) using the provided contract and scripts.
    \item All CSV logs and figures can be reproduced directly from the supplied notebooks and Python scripts.
\end{itemize}

These resources collectively enable complete re-execution of the SlideChain pipeline—from semantic extraction to blockchain anchoring and statistical analysis—ensuring transparent, verifiable, and long-term reproducible results.

\section{SlideChain Smart Contract Source Code}

This appendix provides the complete Solidity implementation of the 
\texttt{SlideChain} smart contract used for all provenance experiments. 
The contract defines the on-chain storage layout, event interface, and 
registration/verification logic for slide-level provenance records. 
It is located at:

\begin{verbatim}
SlideChain/SlideChain/contracts/SlideChain.sol
\end{verbatim}

\noindent The full source code is reproduced below.

\begin{lstlisting}[language=Solidity,basicstyle=\small\ttfamily]
 // SPDX-License-Identifier: MIT
 pragma solidity ^0.8.20;

 /// @title SlideChain - On-chain provenance for MEDI-SLATE slides
 contract SlideChain {
     struct SlideRecord {
         uint256 lectureId;     // 1..23
         uint256 slideId;       // 1..N per lecture
         string slideHash;      // keccak256 hash of SlideX.json as 0x...
         string uri;            // off-chain path (e.g. "Lecture 1/Slide1.json")
         uint256 timestamp;     // block timestamp when registered
         address registrant;    // who registered this slide
     }

     // key = keccak256(lectureId, slideId)
     mapping(bytes32 => SlideRecord) private records;

     event SlideRegistered(
         uint256 indexed lectureId,
         uint256 indexed slideId,
         string slideHash,
         string uri,
         address indexed registrant,
         uint256 timestamp
     );

     /// @notice Compute storage key used for a (lectureId, slideId) pair
     function _key(uint256 lectureId, uint256 slideId)
         internal
         pure
         returns (bytes32)
     {
         return keccak256(abi.encodePacked(lectureId, slideId));
     }

     /// @notice Register a slide hash for a (lectureId, slideId) pair
     /// @dev Reverts if already registered to keep provenance immutable.
     function registerSlide(
         uint256 lectureId,
         uint256 slideId,
         string calldata slideHash,
         string calldata uri
     ) external {
         require(lectureId > 0, "lectureId must be > 0");
         require(slideId > 0, "slideId must be > 0");

         bytes32 key = _key(lectureId, slideId);
         SlideRecord storage existing = records[key];
         require(existing.timestamp == 0, "slide already registered");

         records[key] = SlideRecord({
             lectureId: lectureId,
             slideId: slideId,
             slideHash: slideHash,
             uri: uri,
             timestamp: block.timestamp,
             registrant: msg.sender
         });

         emit SlideRegistered(
             lectureId,
             slideId,
             slideHash,
             uri,
             msg.sender,
             block.timestamp
         );
     }

     /// @notice Get a slide record for (lectureId, slideId)
     function getSlide(uint256 lectureId, uint256 slideId)
         external
         view
         returns (SlideRecord memory)
     {
         bytes32 key = _key(lectureId, slideId);
         return records[key];
     }

     /// @notice Check if a slide is registered
     function isRegistered(uint256 lectureId, uint256 slideId)
         external
         view
         returns (bool)
     {
         bytes32 key = _key(lectureId, slideId);
         return records[key].timestamp != 0;
     }
 }
\end{lstlisting}

\section{Provenance JSON Schema}

\noindent
The following schema defines the canonical structure of SlideChain provenance
records. Each slide produces a single JSON file storing multimodel semantic
extractions, metadata, and file paths. All fields are stable across runs and
serve as the basis for deterministic hashing (keccak256) and blockchain
registration.

\begin{lstlisting}[basicstyle=\small\ttfamily]
{
  "lecture": "string",                // like "Lecture 1"
  "slide_id": "string or integer",    // numeric slide index

  "models": {
    "<model_name>": {                 // for example "InternVL3-14B"
      "concepts": [
        {
          "category": "string",       // "modality", "anatomy"
          "term": "string",           // normalized semantic unit
          "evidence": "string|null"   // optional text snippet
        },
        ...
      ],

      "triples": [
        {
          "s": "string",              // subject
          "p": "string",              // predicate
          "o": "string",              // object
          "confidence": "number|null" // optional float
        },
        ...
      ],

      "evidence": [
        "string",                     // free-text VLM evidence
        ...
      ],

      "raw_output": "string"          // optional, for debugging only
    },

    "... additional VLMs ..."         // total: 4 models
  },

  "paths": {
    "image": "string",                // absolute or relative slide path
    "text": "string",                 // corresponding transcript snippet
    "json": "string"                  // local JSON file path
  },

  "metadata": {
    "timestamp": "string",            // creation time in ISO format
    "source": "string",               // pipeline version or host
    "hash_input_format": "string"     // canonical format used for hashing
  }
}
\end{lstlisting}

\paragraph{Normalization Rules.}
All provenance files follow the same normalization procedure:
\begin{itemize}
    \item concepts and triples are lowercased and whitespace-normalized;
    \item malformed, \texttt{null}, or empty entries are removed;
    \item missing model fields are represented as empty arrays;
    \item JSON keys are serialized with sorted ordering prior to hashing;
    \item evidence fields preserve the exact text returned by each VLM.
\end{itemize}

This schema guarantees that each slide’s semantic representation is stable,
machine-verifiable, and reproducible across extraction runs, enabling the
SlideChain smart contract to detect even a single-character modification in any
provenance file.

\section{Execution Environment}

This section documents the hardware and software environment used to execute all experiments reported in the paper. The purpose is to specify the execution context under which semantic extraction, blockchain registration, and analysis were performed.

All experiments were executed locally on a single workstation. Vision--language model inference, provenance generation, and semantic analysis were implemented in Python. Smart contract deployment, transaction execution, and gas measurement were performed using a local Hardhat Ethereum-compatible development chain.

Tables~\ref{tab:hardware} and~\ref{tab:software} summarize the complete hardware and software configurations used throughout the study.

No external cloud services, third-party inference APIs, or public blockchain networks were used. All measurements—including timestamps, gas usage, and transaction metadata—were collected directly from the local execution environment.

This section is intended to support experimental replication by specifying the execution context. Empirical evaluations of reproducibility, integrity, and semantic stability are reported and analyzed in the main body of the paper.

\end{document}